\documentclass[11pt,a4paper,abstract=on]{scrartcl}

\usepackage[utf8]{inputenc}
\usepackage[english]{babel}

\usepackage[paper=a4paper,left=25mm,right=25mm,top=25mm,bottom=30mm]{geometry}
\usepackage{graphicx}
\usepackage{latexsym}
\usepackage{amsmath,amssymb,amsthm,dsfont}
\usepackage{bbm}
\usepackage{wrapfig}
\usepackage{scrlayer-scrpage}
\usepackage{adjustbox}
\usepackage{tabularx}
\usepackage{mathtools}
\usepackage{booktabs}
\usepackage{bbm}
\usepackage{authblk}
\usepackage{capt-of}
\usepackage{natbib}
\usepackage[breaklinks, colorlinks=true, citecolor=black, urlcolor=black, linkcolor=black]{hyperref}

\usepackage{natbib}
\usepackage{etoolbox}

\makeatletter
\newcommand\bibstyle@comma{\bibpunct(),a,,}
\newcommand\bibstyle@semicolon{\bibpunct();a,,}
\makeatother

\pretocmd\citet{\citestyle{comma}}\relax\relax
\pretocmd\citep{\citestyle{semicolon}}\relax\relax

\numberwithin{equation}{section} 

\newcommand{\R}{\mathbb{R}} 
\newcommand{\N}{\mathbb{N}} 

\newcommand{\code}[1]{\texttt{#1}}

\begin{document}

\title{Machine learning methods for postprocessing ensemble forecasts of wind gusts: \\ A systematic comparison} 


\author[1]{Benedikt Schulz}
\author[1,2]{Sebastian Lerch}
\affil[1]{Karlsruhe Institute of Technology}
\affil[2]{Heidelberg Institute for Theoretical Studies}
\renewcommand\Authands{ and }

\date{\today}

\maketitle

\begin{abstract}
\noindent Postprocessing ensemble weather predictions to correct systematic errors has become a standard practice in research and operations.
However, only few recent studies have focused on ensemble postprocessing of wind gust forecasts, despite its importance for severe weather warnings.
Here, we provide a comprehensive review and systematic comparison of eight statistical and machine learning methods for probabilistic wind gust forecasting via ensemble postprocessing, that can be divided in three groups: 
State of the art postprocessing techniques from statistics (ensemble model output statistics (EMOS), member-by-member postprocessing, isotonic distributional regression), established machine learning methods (gradient-boosting extended EMOS, quantile regression forests) and neural network-based approaches (distributional regression network, Bernstein quantile network, histogram estimation network).
The methods are systematically compared using six years of data from a high-resolution, convection-permitting ensemble prediction system that was run operationally at the German weather service, and hourly observations at 175 surface weather stations in Germany.
While all postprocessing methods yield calibrated forecasts and are able to correct the systematic errors of the raw ensemble predictions, incorporating information from additional meteorological predictor variables beyond wind gusts leads to significant improvements in forecast skill. 
In particular, we propose a flexible framework of locally adaptive neural networks with different probabilistic forecast types as output, which not only significantly outperform all benchmark postprocessing methods but also learn physically consistent relations associated with the diurnal cycle, especially the evening transition of the planetary boundary layer.
\end{abstract}

\section{Introduction} \label{sec:intro}

Wind gusts are among the most significant natural hazards in central Europe. Accurate and reliable forecasts are therefore critically important to issue effective warnings and protect human life and property. However, wind gusts are a challenging meteorological target variable as they are driven by small-scale processes and local occurrence, so that their predictability is limited even for numerical weather prediction (NWP) models run at convection-permitting resolutions. 

In order to quantify forecast uncertainty, most operationally used NWP models generate probabilistic predictions in the form of ensembles of deterministic forecasts that differ in initial conditions, boundary conditions or model specifications. Despite substantial improvements over the past decades \citep{Bauer2015}, ensemble forecasts continue to exhibit systematic errors that require statistical postprocessing to achieve accurate and reliable probabilistic forecasts. Statistical postprocessing has therefore become an integral part of weather forecasting and standard practice in research and operations. We refer to \citet{VannitsemEtAl2018book} for a general introduction to statistical postprocessing and to \citet{Vannitsem2021} for an overview of recent developments.

The focus of our work is on statistical postprocessing of ensemble forecasts of wind gusts. Despite their importance for severe weather warnings, much recent work on ensemble postprocessing has instead focused on temperature, precipitation and mean wind speed. Therefore, our overarching aim is to provide a comprehensive review and systematic comparison of statistical and machine learning methods for ensemble postprocessing specifically tailored to wind gusts.

Most commonly used postprocessing methods are distributional regression approaches where probabilistic forecasts are given by parametric probability distributions with parameters depending on summary statistics of the ensemble predictions of the target variable through suitable link functions. The two most prominent techniques are ensemble model output statistics \citep[EMOS;][]{Gneiting2005}, where the forecast distribution is given by a single parametric distribution, and Bayesian model averaging \citep[BMA;][]{Raftery2005}, where the probabilistic forecast takes the form of a weighted mixture distribution. Both methods tend to perform equally well and the conceptually simpler EMOS approach has become widely used in practice \citep{Vannitsem2021}. In the past decade, a wide range of statistical postprocessing methods has been developed, including physically motivated techniques like member-by-member postprocessing \citep[MBM;][]{VanSchaeybroeck2015} or statistically principled approaches such as isotonic distributional regression \citep[IDR;][]{Henzi2019}.

The use of modern machine learning methods for postprocessing has been the focus of much recent research interest \citep{McGovern2017,Haupt2021,Vannitsem2021}. By enabling the incorporation of additional predictor variables beyond ensemble forecasts of the target variable, these methods have the potential to overcome inherent limitations of traditional approaches. Examples include quantile regression forests \citep[QRF;][]{Taillardat2016} and the gradient-boosting extension of EMOS \citep[EMOS-GB;][]{Messner2017}. \citet{Rasp2018} demonstrate the benefits of a neural network (NN) based distributional regression approach, which we will refer to as distributional regression network (DRN). Their approach is an extension of the EMOS framework that enables nonlinear relationships between arbitrary predictor variables and forecast distribution parameters to be learned in a data-driven way. DRN has been extended towards more flexible, distribution-free approaches based on approximations of the quantile function \citep[Bernstein quantile network, BQN;][]{Bremnes2020} or of quantile-based probabilities that are transformed to a full predictive distribution \citep{Scheuerer2020}.

Many of the postprocessing methods described above have been applied for wind speed prediction, but previous work on wind gusts is scarce. Our work is based on the study of \cite{Pantillon2018}, one of the few exceptions, who use a simple EMOS model for postprocessing to investigate the predictability of wind gusts with a focus on European winter storms.
They find that although postprocessing improves the overall predictive performance, it fails in cases that can be attributed to specific mesoscale structures and corresponding wind gust generation mechanisms. 
As a first step towards the development of more sophisticated methods, we adapt existing as well as novel techniques for statistical postprocessing of wind gusts and conduct a systematic comparison of their predictive performance. 
Our case study utilizes forecasts from the operational ensemble prediction system (EPS) of the German weather service run at convection-permitting resolution for the 2010--2016 period and observations from 175 surface weather stations in Germany. 
In contrast to current research practice, where postprocessing methods are often only compared to a small set of benchmark techniques, our work is the first to systematically compare a wide variety of postprocessing approaches to the best of our knowledge.

The remainder of the paper is structured as follows. Section \ref{sec:data} introduces the data and notation, Section \ref{sec:methods} the various statistical postprocessing methods, whose predictive performance is evaluated in Section \ref{sec:eval}. 
A meteorological interpretation of what the models have learned is presented in Section \ref{ssec:eval_features}. 
Section \ref{sec:conclusions} concludes with a discussion.

\code{R} \citep{RCoreTeam2021} code with implementations of all methods is available at \url{https://github.com/benediktschulz/paper_pp_wind_gusts}.

\section{Data and notation} \label{sec:data}

\subsection{Forecast and observation data} \label{ssec:data_data}

Our study is based on the same dataset as \cite{Pantillon2018} and we refer to their Section 2.1 for a detailed description. The forecasts are generated by the EPS of the Consortium for Small Scale Modelling operational forecast for Germany \citep[COSMO-DE;][]{Baldauf2011COSMO}. The 20-member ensemble is based on initial and boundary conditions from four different global models paired with five sets of physical perturbations and run at a horizontal grid spacing of 2.8 km \citep{Pantillon2018}. In the following, we will refer to the four groups corresponding to the global models, which consist of five members each, as sub-ensembles. 
We consider forecasts that are initialized at 00 UTC with a range from 0 to 21 hours.
The data ranges from December 9, 2010, when the EPS started in pre-operational mode, to the end of 2016, i.e.\ a period of around six years. 

In addition to wind gusts, ensemble forecasts of several other meteorological variables generated by the COSMO-DE-EPS are available. Table \ref{tbl:data_predictors} gives an overview of the 61 meteorological variables as well as additional temporal and spatial predictors derived from station information. The forecasts are evaluated at 175 SYNOP stations in Germany operated by the German weather service, for which hourly observations are available. For the comparison with station data, forecasts from the nearest grid point are taken. 

\begin{table}[t!]
\caption{Overview of available predictors. For the meteorological variables, ensemble forecasts are available, with the term '500--1,000 hPa' referring to the specific model levels at 500, 700, 850, 950 and 1,000 hPa. Spatial predictors contain station-specific information. \label{tbl:data_predictors}}	
\begin{center}
	\scalebox{0.85}{
		\begin{tabular}{lcl}
			\multicolumn{3}{l}{Meteorological variables} \\
			\toprule 
			Feature & Unit & Description \\
			\midrule
			\texttt{VMAX} & m/s & Maximum wind, i.e.\ wind gusts (10m) \\
			\texttt{U} & m/s & U-component of wind (10m, 500--1,000 hPa) \\
			\texttt{V} & m/s & V-component of wind (10m, 500--1,000 hPa) \\
			\texttt{WIND} & m/s & Wind speed, derived from \texttt{U} and \texttt{V} via $\sqrt{\texttt{U}^2 + \texttt{V}^2}$ (10m, 500--1,000 hPa) \\
			\texttt{OMEGA} & Pa/s & Vertical velocity (Pressure) (500--1,000 hPa) \\
			\texttt{T} & K & Temperature (Ground-level, 2m, 500--1,000 hPa) \\
			\texttt{TD} & K & Dew point temperature (2m) \\
			\texttt{RELHUM} & \% & Relative humidity (500--1,000 hPa) \\
			\texttt{TOT\_PREC} & kg/m$^2$ & Total precipitation (Accumulation) \\
			\texttt{RAIN\_GSP} & kg/m$^2$ & Large scale rain (Accumulation) \\
			\texttt{SNOW\_GSP} & kg/m$^2$ & Large scale snowfall - water equivalent (Accumulation) \\
			\texttt{W\_SNOW} & kg/m$^2$ & Snow depth water equivalent \\
			\texttt{W\_SO} & kg/m$^2$ & Column integrated soil moisture (multilayers; 1, 2, 6, 18, 54) \\
			\texttt{CLCT} & \% & Total cloud cover \\
			\texttt{CLCL} & \% & Cloud cover (800 hPa - soil) \\
			\texttt{CLCM} & \% & Cloud cover (400 hPa - 800 hPa) \\
			\texttt{CLCH} & \% & Cloud cover (000 hPa - 400 hPa) \\
			\texttt{HBAS\_SC} & m & Cloud base above mean sea level, shallow connection \\
			\texttt{HTOP\_SC} & m & Cloud top above mean sea level, shallow connection \\
			\texttt{ASOB\_S} & W/m$^2$ & Net short wave radiation flux (at the surface) \\
			\texttt{ATHB\_S} & W/m$^2$ & Net long wave radiation flux (m) (at the surface) \\
			\texttt{ALB\_RAD} & \% & Albedo (in short-wave) \\
			\texttt{PMSL} & Pa & Pressure reduced to mean sea level \\
			\texttt{FI} & m$^2$/s$^2$ & Geopotential (500--1,000 hPa) \\
			\midrule \\[0.5em]
			\multicolumn{3}{l}{Other predictors} \\
			\midrule
			Feature & Type & Description \\
			\midrule
			\texttt{yday} & Temporal & Cosine transformed day of the year: $\cos \left(2\pi \frac{t - 1}{365} \right)$, where $t$ is the day of the year. \\
			\texttt{lat} & Spatial & Latitude of the station. \\
			\texttt{lon} & Spatial & Longitude of the station. \\
			\texttt{alt} & Spatial & Altitude of the station. \\
			\texttt{orog} & Spatial & Difference of station altitude and model surface height of nearest grid point. \\
			\texttt{loc\_bias} & Spatial & Mean bias of wind gust ensemble forecasts from 2010--2015 at the station. \\
			\texttt{loc\_cover} & Spatial & Mean coverage of the wind gust ensemble forecasts from 2010--2015 at the station. \\
			\bottomrule
		\end{tabular}
	}
\end{center}
\end{table}

\subsection{Training, testing and validation} \label{ssec:data_split}

The postprocessing methods that will be presented in Section \ref{sec:methods} are trained on a set of past forecast-observation pairs in order to correct the systematic errors of the ensemble predictions. Even though many studies are based on rolling training windows consisting of the most recent days only, we will use a static training period. This is common practice in the operational use of postprocessing models \citep{Hess2020} and can be motivated by studies suggesting that using long archives of training data often lead to superior performance, irrespective of potential changes in the underlying NWP model or the meteorological conditions \citep{Lang2020}.

Therefore, we will use the period of 2010--2015 as training set and 2016 as independent test set.
The implementation of most of the methods requires the choice of a model architecture and the tuning of specific hyperparameters. 
To avoid overfitting in the model selection process, we further split the training set into the period of 2010--2014 which is used for training, and use the year 2015 for validation purposes. After finalizing the choice of the most suitable model variant based on the validation period, the entire training period from 2010--2015 is used to fit that model for the final evaluation on the test set.

\subsection{Notation} \label{ssec:data_notation}

Next, we will briefly introduce the notation used in the following. The weather variable of interest, i.e.\ the speed of wind gusts, will be denoted by $y$ and ensemble forecasts of wind gusts by $x$. Note that $y > 0$, and that $x = (x_1,\dots,x_m)$ is an $m$-dimensional vector, where $m$ is the ensemble size and $x_i$ the $i$-th ensemble member. The ensemble mean of $x$ is denoted by $\overline{x}$ and the standard deviation by $s(x)$. 

We will use the term predictor or feature exchangeably to denote a predictor variable that is used as an input to a postprocessing model. For most meteorological variables, we will typically use the ensemble mean as predictor. If a method uses several predictors, we will refer to the vector including all predictors by $\pmb{x} \in \R^p$, where $p$ is the number of predictors and $\pmb{x}_i$ is the $i$-th predictor. For a set of past observations, ensemble forecasts of wind gusts and predictors, we will denote the variables by $y_j$, $x_{\cdot j}$ and $\pmb{x}_{\cdot j}$, where $j = 1, \dots, n$ and $n$ is the training set size.

\section{Postprocessing methods} \label{sec:methods}

This section introduces the postprocessing methods that are systematically compared for ensemble forecasts of wind gusts. The postprocessing methods will be presented in three groups, starting with established, comparatively simple techniques rooted in statistics. We will proceed to machine learning approaches ranging from random forest and gradient boosting techniques up to methods based on NNs. A full description of all implementation details is deferred to Appendix \ref{sec:appendix_implementation}. Note that for each of the postprocessing methods, we fit a separate model for each lead time based on training data only consisting of cases corresponding to that lead time.

\subsection{Basic techniques} \label{ssec:methods_basic}

As a first group of methods, we will review basic statistical postprocessing techniques, where the term basic refers to the fact that these methods solely use the ensemble forecasts of wind gusts as predictors. They are straightforward to implement and serve as benchmark methods for the more advanced approaches.

\subsubsection{Ensemble Model Output Statistics (EMOS)} \label{sssec:methods_emos}

EMOS, originally proposed by \cite{Gneiting2005} and sometimes referred to as nonhomogeneous regression, is one of the most prominent statistical postprocessing methods. EMOS is a distributional regression approach, which assumes that, given an ensemble forecast $x$, the weather variable of interest $Y$ follows a parametric distribution $\mathcal{L} \left( \theta \right)$ with $\theta \in \Theta$, where $\Theta$ denotes the parameter space of $\mathcal{L}$. The distribution parameter (vector) $\theta$ is connected to the ensemble forecast via a link function $g$ such that
\begin{eqnarray}
	Y \mid x \sim \mathcal{L} \left( \theta \right), \quad \theta = g \left( x \right) \in \Theta. \label{eq:emos_distribution}
\end{eqnarray}
The choice of the parametric family for the forecast distribution depends on the weather variable of interest. \cite{Gneiting2005} use a Gaussian distribution for temperature and sea level pressure forecasts. More complex variables variables like precipitation or solar irradiance have been modelled via zero-censored distributions, whose mixed discrete-continuous nature enables point masses for the events of no rain or no irradiance \citep[see, e.g.,][]{Scheuerer2014,Schulz2021}. In contrast to these variables, wind speed is assumed to be strictly positive and modeled via distributions that are left-truncated at zero. In the extant literature, positive distributions that have been employed include truncated normal \citep{Thorarinsdottir2010}, truncated logistic \citep{Messner2014}, log-normal \citep{Baran2015} or truncated generalized extreme value \citep[GEV;][]{Baran2021} distributions. While the differences observed in terms of the predictive performance for different distributional models are generally only minor, combinations or weighted mixtures of several parametric families have been demonstrated to improve calibration and forecast performance for extreme events \citep{Lerch2013, BaranLerch2016, Baran2018}.

While the parametric families employed for wind speed can be assumed to be appropriate to model wind gusts as well, specific studies tailored to wind gusts are scarce. We implemented the truncated versions of the logistic, normal and GEV distribution and found only small differences between the models in agreement with \cite{Pantillon2018}, and use a truncated logistic distribution in the following. Given a logistic distribution with cumulative distribution function (CDF)
\begin{eqnarray*}
	F \left(z; \mu, \sigma\right) :=  \left( 1 + \exp \left( - \frac{z - \mu}{\sigma} \right) \right)^{-1}, \quad z \in \R, 
\end{eqnarray*}
and probability density function (PDF)
\begin{eqnarray}
	f \left(z; \mu, \sigma\right) :=  \dfrac{ \exp \left(- \frac{z - \mu}{\sigma}\right) }{ \sigma \left( 1 + \exp \left(- \frac{z - \mu}{\sigma}\right) \right)^2 }, \quad z \in \R. \label{eq:logis_pdf}
\end{eqnarray}
where $\mu$ is the location and $\sigma > 0$ the scale parameter, the logistic distribution left-truncated in zero is given by the CDF
\begin{eqnarray*}
	F_0 \left(z; \mu, \sigma\right) := \dfrac{F \left(z; \mu, \sigma\right) - F \left(0; \mu, \sigma\right)}{1 - F \left(0; \mu, \sigma\right)}, \quad z > 0. 
\end{eqnarray*}
Note that, in contrast to the logistic distribution, the mean of the truncated logistic distribution is not equivalent to the location parameter $\mu$, which may still be negative. 



In our EMOS model, the distribution parameters $\mu$ and $\sigma$ are linked to the ensemble forecast $x$ via 
\begin{eqnarray}
	\mu \left(x; a, b \right) := a + b \cdot \overline{x}, \quad \sigma \left(x; c, d \right) := \exp \left(c + d \cdot \log \left(s(x)\right) \right), 
	\label{eq:emos_links}
\end{eqnarray}
where $a, c, d \in \R$ and $b > 0$ are estimated via optimum score estimation, i.e.\ by minimizing a strictly proper scoring rule \citep{Gneiting2007scoring}. For details on forecast evaluation, see Appendix \ref{sec:appendix_eval}.
We here estimate the parameters by minimizing the continuous ranked probability score (CRPS), for which we observed similar results to maximum likelihood estimation (MLE). Analytical expressions of the CRPS and the corresponding gradient function of a truncated logistic distribution are available in the \texttt{scoringRules} package \citep{Jordan2019scoringrules}. 
Note that we do not specifically account for the existence of sub-ensembles in \eqref{eq:emos_links}, since initial experiments suggested a degradation of predictive performance.

The EMOS coefficients $a,b,c,d$ are estimated locally, i.e.\ we estimate a separate model for each station in order to account for station-specific error characteristics of the ensemble predictions. 
In addition, we employ a seasonal training scheme where a training set consists of all forecast cases of the previous, current and next month with respect to (w.r.t.) the date of interest. This results in twelve different training sets for each station, one for each month, that enable an adaption to seasonal changes. 
In accordance with the results in \citet{Lang2020}, this seasonal approach outperforms both a rolling training window as well as training on the entire set. 

\subsubsection{Member-By-Member Postprocessing (MBM)} \label{sssec:methods_mbm}

MBM \citep{VanSchaeybroeck2015} is based on the idea to adjust each member individually in order to generate a calibrated ensemble. \cite{Wilks2018chapter} highlights several variants of MBM, which are of the general form
\begin{eqnarray}
	\tilde{x}_i := (a + b \cdot \overline{x}) + \gamma \cdot \left( x_i - \overline{x}\right), \quad i = 1, \dots, m, \quad a, b, \gamma \in \R, \label{eq:mbm_general}
\end{eqnarray}
where $\tilde{x} = (\tilde{x}_1, \dots, \tilde{x}_m)$ denotes the postprocessed ensemble. The first term in \eqref{eq:mbm_general} represents a bias-corrected ensemble mean, whereas the second term includes the individual members with $\gamma$ scaling the distance to the ensemble mean. 
Our implementation of MBM postprocessing follows \cite{VanSchaeybroeck2015}, who let the stretch coefficient $\gamma$ depend on the ensemble mean difference $\delta$ via
\begin{eqnarray}
	\gamma := c + \dfrac{d}{\delta (x)}, \quad \text{where} \quad \delta (x) := \dfrac{1}{m^2} \sum_{i,l = 1}^{m} \left| x_i - x_l \right| \quad \text{and} \quad c, d \in \R. \label{eq:mbm_gamma}
\end{eqnarray}

The MBM parameters $a,b,c,d$ are estimated by minimizing the CRPS of the adjusted ensemble, the loss function thus corresponds to 
\begin{eqnarray*}
    l \left( a, b, c, d \right) &:=& \dfrac{1}{n} \sum_{j = 1}^{n} \left[ \left( \dfrac{1}{m} \sum_{i = 1}^{m} \left| \tilde{x}_{ij} - y_j \right| \right) - \dfrac{\delta (\tilde{x}_{\cdot j})}{2} \right]. 
\end{eqnarray*}
Note that the calibrated ensemble $\tilde{x}$ depends on the MBM parameters via \eqref{eq:mbm_general} and \eqref{eq:mbm_gamma}. The accuracy of each ensemble member is penalized in the first term, the second term penalizes the sharpness via the ensemble mean difference. \cite{VanSchaeybroeck2015} further consider a MLE approach under a parametric assumption, which, however, resulted in less well calibrated forecasts with slightly worse overall performance.

Due to the fact that the existence of sub-ensembles results in systematic deviations from calibration, especially for a lead time of 0 hours, we extend MBM to account for the sub-ensembles generated by the different models. Illustrations and additional results are provided in the supplemental material.
By introducing a function $k(i) = \left \lceil{i /5}\right \rceil \in \lbrace 1, \dots, 4 \rbrace$ for $i = 1, \dots, 20$, that identifies the sub-ensemble to which the $i$-th member belongs, we can incorporate the sub-model structure by modifying \eqref{eq:mbm_general} to
\begin{eqnarray*}
	\tilde{x}_i := (a + b_{k(i)} \cdot \overline{x}_{k(i)}) + \left( c + \dfrac{d_{k(i)}}{\delta_{k(i)} (x)} \right) \cdot \left( x_i - \overline{x}_{k(i)}\right), \quad i = 1, \dots, 20, 
\end{eqnarray*}
where $\overline{x}_k$ denotes the mean of the $k$-th sub-ensemble in $x$, $\delta_k (x)$ denotes the mean difference of the $k$-th sub-ensemble in $x$, $k = 1, \dots, 4$, and $a, b_1, \dots, b_4, c, d_1, \dots, d_4 \in \R$ are the MBM parameters. This increases the number of parameters by six, but substantially improves performance and mostly eliminates the relicts of the sub-ensemble structure. One downside of this modification is that it increases the computational time drastically, here by a factor of around 17. 
As alternative modifications, we also applied MBM separately to each sub-ensemble and use a separate parameter $c$ for each sub-model, which performs equally well but is more complex. 

The training is performed analogously to EMOS, utilizing a local and seasonal training scheme. In particular, accounting for potential seasonal changes via seasonal training substantially improved performance compared to using the entire available training set. A main advantage of MBM compared to all other approaches is that the rank correlation structure of the ensemble forecasts is preserved by postprocessing, since each member is transformed individually by the same linear transformation. MBM thus results in forecasts that are physically consistent over time, space and also different weather variables, even if MBM is applied for each component separately \citep{VanSchaeybroeck2015,Schefzik2017,Wilks2018chapter}.

\subsubsection{Isotonic Distributional Regression (IDR)} \label{sssec:methods_idr}

\cite{Henzi2019} propose IDR, a novel non-parametric regression technique which results in simple and flexible probabilistic forecasts as it depends neither on distributional assumptions nor pre-specified transformations.
Since it requires no parameter tuning and minimal implementation choices, it is an ideal generic benchmark in probabilistic forecasting tasks.
IDR is built on the assumption of an isotonic relationship between the predictors and the target variable. In the univariate case with only one predictor, isotonicity is based on the linear ordering on the real line. When multiple predictors (such as multiple ensemble members) are given, the multivariate covariate space is equipped with a partial order. Under those order restrictions, a conditional distribution that is optimal w.r.t.\ a broad class of relevant loss functions including proper scoring rules is then estimated. Conceptually, IDR can be seen as a far-reaching generalization of widely used isotonic regression techniques that are based on the pool-adjacent-violators algorithm \citep{DeLeeuw2009}. To the best of our knowledge, our work is the first application of IDR in a postprocessing context besides the case study on precipitation accumulation in \citet{Henzi2019} who find that IDR forecasts were competitive to EMOS and BMA.

The only implementation choice required for IDR is the selection of a partial order on the covariate space. Among the choices introduced in \citet{Henzi2019}, the empirical stochastic order (SD) and the empirical increasing convex order (ICX) are appropriate for the situation at hand when all ensemble members are used as predictors for the IDR model. We selected SD which resulted in slightly better results on the validation data. We further considered an alternative model formulation where only the ensemble mean was used as predictor, which reduces to the special case of a less complex distributional (single) index model \citep{Henzi2020}, but did not improve predictive performance.

We implement IDR as a local model, treating each station separately since it is not obvious how to incorporate station-specific information into the model formulation. Given the limited amount of training data available, we further consider only the wind gust ensemble as predictor variable. Following suggestions of \citet{Henzi2019}, we use subsample aggregation (subbagging) and apply IDR on 100 random subsamples half the size of the available training set. IDR is implemented using the \code{isodistrreg} package \citep{Henzi2019isodistrreg}.

\subsection{Incorporating additional information via machine learning methods} \label{ssec:methods_ml}

The second group of postprocessing methods consists of machine learning methods that are able to incorporate additional predictor variables besides ensemble forecasts of wind gusts. As noted in Section \ref{sec:data}, ensemble forecasts of 60 additional variables are available. Including the additional information into the basic approaches introduced in Section \ref{ssec:methods_basic} is in principle possible, but far from straightforward since one has to carefully select appropriate features and take measures to prevent overfitting. By contrast, the machine learning approaches presented here offer a more feasible approach to include additional predictors in an automated, data-driven way.

\subsubsection{Gradient-Boosting Extension of EMOS (EMOS-GB)} \label{sssec:methods_emos_gb}

\citet{Messner2017} propose an extension of EMOS which allows for including additional predictors by using link functions that depend on $\pmb{x}$ instead of $x$ in \eqref{eq:emos_distribution},
\begin{align}
    \mu \left(\pmb{x}; a, b_1, \dots, b_p \right) &:= a + \sum_{i = 1}^{p} b_i \cdot \pmb{x}_i,
	\quad a, b_1, \dots, b_p \in \R,\\ 
	\sigma \left(\pmb{x}; c, d_1, \dots, d_p \right) &:= \exp \left( c + \sum_{i = 1}^{p} d_i \cdot \pmb{x}_i \right), 
	\quad c, d_1, \dots, d_p \in \R.
	\label{eq:emos_gb_links}
\end{align}
To enable an automated selection of relevant predictors, \cite{Messner2017} follow a gradient boosting approach. Given a large set of predictors, the idea of boosting is to initialize all coefficient values at zero, and to iteratively update only those coefficients corresponding to the predictor that improves the predictive performance most.  Based on the gradient of the loss function, the predictor with the highest correlation to the gradient is selected and then the corresponding coefficient is updated by taking a step in direction of the steepest descent of the gradient. This procedure is repeated until a stopping criterion is reached to avoid overfitting.

To ensure comparability with the basic EMOS approach, we employ a truncated logistic distribution for the probabilistic forecasts. The parameters are determined using MLE which resulted in superior predictive performance based on initial tests on the validation data compared to minimum CRPS estimation. 
We use the ensemble mean and standard deviation of all meteorological variables in Table \ref{tbl:data_predictors} as inputs to the EMOS-GB model. Note that in contrast to the other advanced postprocessing methods introduced below, we here include the standard deviation of all variables as potential predictors since we found this to improve the predictive performance.
Further, we include the cosine-transformed day of the year in order to adapt to seasonal changes, since seasonal training approaches as applied for EMOS and MBM lead to numerically unstable estimation procedures and degraded forecast performance. 
Although spatial predictors can in principle be included in a similar fashion, we estimate EMOS-GB models locally since we found this approach to outperform a joint model for all stations by a large margin.
Our implementation of EMOS-GB is based on the \code{crch} package \citep{Messner2016crch}.

\subsubsection{Quantile Regression Forest (QRF)} \label{sssec:methods_qrf}

A non-parametric, data-driven technique that neither relies on distributional assumptions, link functions nor parameter estimation is QRF, which was first used in the context of postprocessing by \cite{Taillardat2016}. 
Random forests are randomized ensembles of decision trees, which operate by splitting the predictor space in order to create an analog forecast \citep{Breiman1984}. 
This is done iteratively by first finding an order criterion based on the predictor that explains the variability within the training set best, and then splitting the predictor space according to this criterion. This procedure is repeated on the resulting subsets until convergence is reached, thereby creating a partition of the predictor space. Following the decisions at the so-called nodes, one then obtains an analog forecast based on the training samples. 
Random forests create an ensemble of decision trees by considering only a randomly chosen subset of the training data at each tree and of the predictors at each node, aiming to reduce correlation between individual decision trees \citep{Breiman2001}.  
QRF extends the random forest framework by performing a quantile regression that generates a probabilistic forecast \citep{Meinshausen2006}. The QRF forecast thus approximates the forecast distribution by a set of quantile forecasts derived from the set of analog observations.

In contrast to EMOS-GB, only the ensemble mean values of the additional meteorological variables are integrated as predictor variables, since we found that including the standard deviations as well led to more overdispersed forecasts and degraded forecast performance. A potential reason is given by the random selection of predictors at each node, which limits the automated selection of relevant predictors in that a decision based on a subset of irrelevant predictors only may lead to overfitting \citep[Section 15.3.4]{Hastie2009}.

Although spatial predictors can be incorporated into a global, joint QRF model for all stations generating calibrated forecasts, we found that the extant practice of implementing local QRF models separately at each station \citep{Taillardat2016,Rasp2018} results in superior predictive performance and avoids the increased computational demand both in terms of required calculations and memory of a global QRF variant \citep{Taillardat2020}.
Our implementation is based on the \code{ranger} package \citep{Wright2017ranger}.

\subsection{Neural network-based methods} \label{ssec:methods_nn}

Over the past decade, NNs have become ubiquitous in data-driven scientific disciplines and have in recent years been increasingly used in the postprocessing literature \citep[see, e.g.][for a recent review]{Vannitsem2021}. 
NNs are universal function approximators for which a variety of highly complex extensions has been proposed. However, NN models often require large datasets and computational efforts, and are sometimes perceived to lack interpretability.

In the following, we will first present a third group of postprocessing methods based on NNs. Following the introduction of a general framework of our network-based postprocessing methods, we will introduce three model variants and discuss how to combine an ensemble of networks.
In the interest of brevity, we will assume a basic familiarity with NNs and the underlying terminology. We refer to \cite{McGovern2019} for an accessible introduction in a meteorological context and to \cite{Goodfellow2016} for a detailed review. 

\subsubsection{A framework for neural network-based postprocessing}

\begin{figure}
\begin{center}
\includegraphics[width=\textwidth]{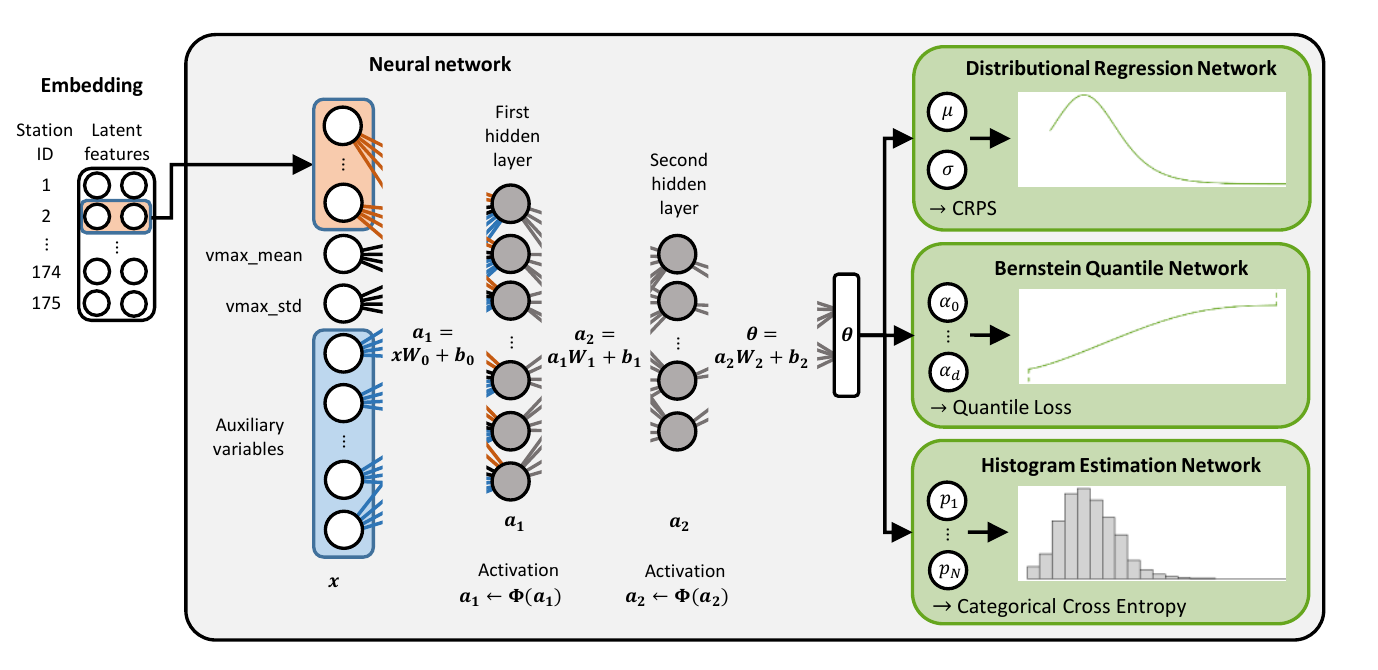}
\caption{Graphical illustration of the framework for neural network-based postprocessing presented in Section \ref{ssec:methods_nn}. \label{fig:nn_framework}}
\end{center}
\end{figure}

The use of NNs in a distributional regression-based postprocessing context was first proposed in \citet{Rasp2018}. Our framework for NN-based postprocessing builds on their approach and subsequent developments in \cite{Bremnes2020, Scheuerer2020}, and \citet{Veldkamp2021}, among others. In particular, we propose three model variants that utilize a common basic NN architecture, but differ in terms of the form that the probabilistic forecasts take which governs both the output of the NN as well as the loss function used for parameter estimation. A graphical illustration of this framework is presented in Figure \ref{fig:nn_framework}.

The rise of artificial intelligence and NNs is closely connected to the increase in data availability and computing power, as these methods unfold their strengths when modelling complex nonlinear relations trained on large datasets. A main challenge in the case of postprocessing is to find a way to optimally utilize the entirety of available input data while preserving the inherent spatial and temporal information. We focus on building one network jointly for all stations at a given lead time, which we will refer to as locally adaptive joint network. For this purpose \cite{Rasp2018} propose a station embedding, where a station identifier is mapped to a vector of latent features, which are then used as auxiliary input variables of the NN. The estimation of the embedding mapping is integrated into the overall training procedure and aims to implicitly model local characteristics. 

Our basic NN architecture consists of two hidden layers and a customized output. The training procedure is based on the adaptive moment estimation (Adam) algorithm \citep{Kingma2015}, and the weights of the network are estimated on the training period of 2010--2014 by optimizing a suitable loss function tailored to the desired output. We apply an early-stopping strategy that stops the training process when the validation loss remains constant for a given number of epochs to prevent the model from overfitting. 

To determine hyperparameters of the NN models such as the learning rate, the embedding dimension or the number of nodes in a hidden layer, we perform a semi-automated hyperparameter tuning based on the validation set. Overall, the results are relatively robust to a wide range of tuning parameter choices, and we found that increasing the number of layers or the number of nodes in a layer did not improve predictive performance. Compared to the models used in \cite{Rasp2018}, we increased the embedding dimension and used a softplus activation function. The exact configuration slightly varies across the three model variants introduced in the following. In addition to the station embedding, the spatial features in Table \ref{tbl:data_predictors} and the temporal predictor, we found that including only the mean values of the meteorological predictors, but not the corresponding standard deviations, improved the predictive performance. These results are in line with those of QRF and those of \cite{Rasp2018} who find that the standard deviations are only of minor importance for explaining and improving the NNs predictions.

To account for the inherent uncertainty in the training of NN models via stochastic gradient descent methods and to improve predictive performance, we again follow \cite{Rasp2018} and create an ensemble of ten networks for each variant which we aggregate into a final forecast. The combination of the corresponding predictive distributions is discussed in Section \ref{sssec:nn_combining} following the description of the three model variants. All NN models were implemented via the \code{R} interface to \code{keras} \citep[2.4.3;][]{Allaire2020keras} built on \code{tensorflow} \citep[2.3.0;][]{Allaire2020tf}.

\subsubsection{Distributional Regression Network (DRN)} \label{sssec:methods_drn}

\cite{Rasp2018} propose a postprocessing method based on NNs that extends the EMOS framework. A key component of the improvement is that instead of relying on pre-specified link function such as \eqref{eq:emos_links} or \eqref{eq:emos_gb_links} to connect input predictors to distribution parameters, a NN is used to automatically learn flexible, non-linear relations in a data-driven way. 
The output of the NN is thus given by the forecast distribution parameters, and the weights of the network are learned by optimizing the CRPS, which performed equally well to MLE.

We adapt DRN to wind gust forecasting by using a truncated logistic distribution. In contrast to the Gaussian predictive distribution used by \cite{Rasp2018}, this leads to additional technical challenges due to the truncation, i.e.\ the division by $1 - F (0; \mu, \sigma)$, which induces numerical instabilities. To stabilize training, we enforce $\mu \geq 0$ by applying a softplus activation function in the output nodes, resulting in $1 - F (0; \mu, \sigma) \geq 0.5$. 
Note that the mode of the truncated logistic distribution is given by $\max(\mu, 0)$ and thus the reduction in flexibility can be seen as a restriction to positive modes. 
Given that negative location parameters tend to be only estimated for wind gusts of low-intensity, which are in general of little interest, and that we noticed no effect on the predictive performance, the restrictions of the parameter space can be considered to be negligible.

\subsubsection{Bernstein Quantile Network (BQN)} \label{sssec:methods_bqn}

\cite{Bremnes2020} extends the DRN framework of \citet{Rasp2018} towards a nonparametric approach where the quantile function of the forecast distribution is modeled by a linear combination of Bernstein polynomials. BQN was proposed for wind speed forecasting, but can be readily applied to other target variables of interest due to its flexibility. The associated forecast is given by the quantile function
\begin{eqnarray}
	Q \left( \tau \mid \pmb{x} \right) := \sum_{l = 0}^{d} \alpha_l (\pmb{x}) B_{l,d} (\tau), \quad \tau \in \left[0, 1\right], \label{eq:bqn_quantile_function}
\end{eqnarray}
with basis coefficients $0 \leq \alpha_0 (\pmb{x}) \leq \dots \leq \alpha_d (\pmb{x})$, where
\begin{eqnarray*}
	B_{l,d} (\tau) = \binom{d}{l} \tau^l \left( 1 - \tau \right)^{d - l}, \quad l = 0,1, \dots, d, 
\end{eqnarray*}
are the Bernstein basis polynomials of degree $d \in \N$. The key implementation choice is the degree $d$, with larger values leading to a more flexible forecast but also a larger estimation variance.

For a given degree $d$, the BQN forecast is fully defined by the $d+1$ basis coefficients, which are linked to the predictors via a NN. A critical requirement is that the coefficients are non-decreasing, because this implies that the quantile function is also non-decreasing\footnote{If the coefficients are strictly increasing, the same holds for the quantile function: Omitting the conditioning on $\pmb{x}$, the derivative of \eqref{eq:bqn_quantile_function} is given by $Q' (\tau) = d \sum_{l = 0}^{d-1} ( \alpha_{l+1} - \alpha_l) B_{l,d-1} (\tau)$ \citep{Wang2012Bernstein}. The derivative is positive for $\tau \in (0, 1)$ if the coefficients are strictly increasing, because the Bernstein polynomials are also positive on the open unit interval.}. 
Further, positive coefficients result in a positive support of the forecast distribution.
In contrast to \cite{Bremnes2020}, we enforce these conditions by targeting the increments $\tilde{\alpha}_l \geq 0$, $l = 0, \dots, d$, based on which the coefficients can be derived via the recursive formula
\begin{eqnarray*}
    \alpha_0 = \tilde{\alpha}_0, \quad \alpha_l = \alpha_{l - 1} + \tilde{\alpha}_l, \quad l = 1, \dots, d. 
\end{eqnarray*}
We obtain the increments as output of the NN and apply a softplus activation function in the output layer, which ensures positivity of the increments and thereby strictly increasing quantile functions, and improves the predictive performance.

Due to the lack of a readily available closed form expression of the CRPS or LS of BQN forecasts and following \citet{Bremnes2020}, the network parameters are estimated based on the quantile loss (QL), which is a consistent scoring function for the corresponding quantile and integration over the unit interval is proportional to the CRPS \citep{Gneiting2011Ranjan}. 
We average the QL over 99 equidistant quantiles corresponding to steps of 1\%. Regarding the degree of the Bernstein polynomials, \citet{Bremnes2020} considers a degree of 8. We found that increasing the degree to 12 resulted in better calibrated forecasts and improved predictive performance on the validation data. Again following \citet{Bremnes2020} and in contrast to our implementation of DRN and the third variant introduced below, we use all 20 ensemble member predictions of wind gust as input instead of the ensemble mean and standard deviation.

\subsubsection{Histogram Estimation Network (HEN)} \label{sssec:methods_hen}

The third network-based postprocessing method may be considered as a universally applicable approach to probabilistic forecasting and is based on the idea to transform the probabilistic forecasting problem into a classification task, one of the main applications of NNs.
This is done by partitioning the observation range in distinct classes and assigning a probability to each of them. 
In mathematical terms, this is equivalent to assuming that the probabilistic forecast is given by a piecewise uniform distribution. The PDF of such a distribution is given by a piecewise constant function, which resembles a histogram, we thus refer to this approach as histogram estimation network (HEN). Variants of this approach have been used in a variety of disciplines and applications \citep[e.g.,][]{Felder2018,Gasthaus2019,Li2021}. For recent examples in the context of postprocessing, see \citet{Scheuerer2020} and \citet{Veldkamp2021}.

To formally introduce the HEN forecasts, let $N$ be the number of bins and $b_0 < \dots < b_N$ the edges of the bins $I_l = [b_{l-1}, b_l)$ with probabilities $p_l$, $l = 1, \dots, N$, where it holds that $\sum_{l = 1}^{N} p_l = 1$. 
For simplicity, we assume that the observation $y$ falls within the bin range, i.e.\ $b_0 \leq y < b_N$, and define the bin in which the observation falls via
\begin{eqnarray*}
	k := k(y) 
	:= \max \lbrace l: b_l \leq y,\, 1 \leq l \leq N \rbrace.
\end{eqnarray*}

The PDF of the corresponding probabilistic forecast is a piecewise constant function given by
\begin{eqnarray*}
	f (y) = \sum_{l = 1}^{N} \dfrac{p_l}{b_l - b_{l-1}} \cdot \mathbbm{1} \lbrace y \in I_l \rbrace = \dfrac{p_k}{b_k - b_{k-1}}, 
	\quad b_0 \leq y < b_N. 
\end{eqnarray*}

The associated CDF is a piecewise linear function given by
\begin{eqnarray}
	F (y) = p_k \dfrac{ y - b_{k-1} }{ b_k - b_{k-1} } + \sum_{l = 1}^{k - 1} p_l, 
	\quad b_0 \leq y < b_N, \label{eq:hen_cdf}
\end{eqnarray}
the piecewise linear quantile function is defined in a similar manner. In contrast to the CDF, the edges of the quantile function are given by the accumulated bin probabilities and not $b_0, \dots, b_N$. Note that the HEN forecast is defined by the binning scheme and the corresponding bin probabilities.


The simple structure of the HEN forecast allows for computing analytical solutions of the CRPS and LS that can be used for parameter estimation. The LS is given by
\begin{eqnarray}
	\text{LS} (f, y) = - \log \dfrac{p_k}{b_k - b_{k-1}} = \log \left( b_k - b_{k-1} \right) - \log \left( p_k \right), \label{eq:hen_logs}
\end{eqnarray}
a corresponding formula for the CRPS is provided in the supplemental material.

Given a fixed number of bins specified as a hyperparameter, the bin edges and corresponding bin probabilities need to be determined. There exist several options for the output of the NN architecture to achieve this. The most flexible approach, for example implemented in \citet{Gasthaus2019}, would be to obtain both the bin edges and the probabilities as output of the NN. We here instead follow a more parsimonious alternative and fix the bin edges, so that only the bin probabilities are determined by the NN, which can be interpreted as a probabilistic classification task.\footnote{
Note that alternatively, it is also possible to fix the bin probabilities and determine the bin edges by the NN. 
This would be equivalent to estimating the quantiles of the forecast distribution at the levels defined by the prespecified probabilities, i.e.\ a quantile regression.
} 
Minimizing the LS in \eqref{eq:hen_logs}, then reduces to minimizing $\tilde{\text{LS}} (f, y) = - \log \left( p_k \right)$, which is referred to as categorical cross-entropy in the machine learning literature. 
Thus, one can estimate the HEN forecast via a standard classification network preserving the optimum scoring principle.

For our application to wind gusts, we found that the binning scheme is an essential factor, e.g.\ a fine equidistant binning of length 0.5 m/s leads to physically inconsistent forecasts. Based on initial experiments on the validation data, we devise a data-driven binning scheme starting from one bin per unique observed value and merging bins to end up with a total number of 20. Details on this procedure are provided in Appendix \ref{sec:appendix_implementation}. We apply a softmax function to the NN output nodes to ensure that the obtained probabilities sum to 1. The network parameters are estimated using the categorical cross-entropy.

\subsubsection{Combining predictive distributions}\label{sssec:nn_combining}

NN-based forecasting models are often run several times from randomly initialized parameter values to produce an ensemble of predictions in order to account for the randomness of the training process based on stochastic gradient descent methods. 
We follow this principle and produce an ensemble of ten models for each of the three variants of NN-based postprocessing models introduced above, which leads to the question how the resulting ensembles of NN outputs should be aggregated into a single probabilistic forecast for every model. We here briefly discuss this aggregation step.

For DRN, the output of the ensemble of NN models is a collection of pairs of location and scale parameters of the forecast distributions. Following \cite{Rasp2018}, we aggregate these forecast by averaging the distribution parameters. 
To aggregate the ensemble of BQN predictions given by sets of coefficient values of the Bernstein polynomials, we follow \citet{Bremnes2020} and average the individual coefficient values across the multiple runs. This is equivalent to an equally weighted averaging of quantile functions, also referred to as Vincentization \citep{Genest1992}.

\begin{figure}
\begin{center}
		\includegraphics[width=\textwidth]{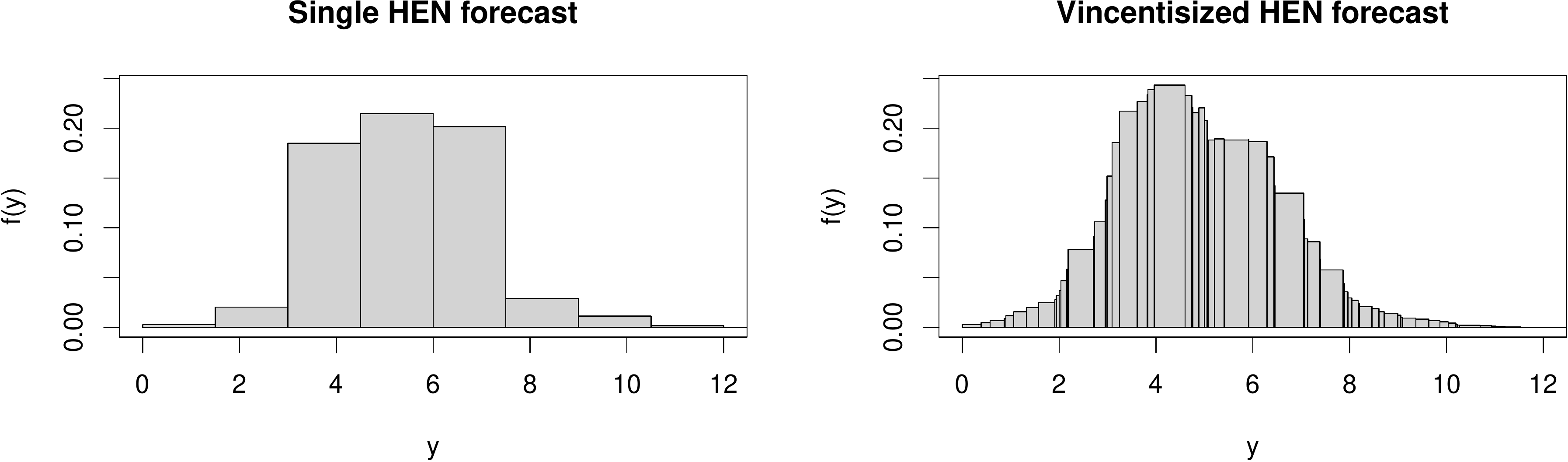}
		\caption{Predictive PDF of a single HEN forecast (left) and the corresponding aggregated forecast where the single forecast is combined with nine other HEN model runs via Vincentization (right). \label{fig:hen_vincentization}}
\end{center}
\end{figure}

In the case of an ensemble of HEN models, the output is given by sets of bin probabilities. Instead of simply averaging the bin probabilities across the ensemble of model runs, which is known to yield underconfident forecasts that lack sharpness \citep{Ranjan2010,Gneiting2013}, we take a Vincentization approach. Since the quantile function is a piecewise linear function with edges depending on the bin probabilities, the predictions of the individual networks are subject to a different binning w.r.t.\ the quantile function. The average of those piecewise linear functions is again a piecewise linear function, where the edges are given by the union of all individual edges. This procedure leads to a smoothed final probabilistic forecast with a much finer binning compared to the individual model runs, and eliminates the downside of fixed bin edges that may be too coarse and not capable of flexibly adapting to current weather conditions. An illustration of this effect is provided in Figure \ref{fig:hen_vincentization}.

\subsection{Summary and comparison of postprocessing methods}

\begin{figure}[p]
\begin{center}
\includegraphics[width=0.75\textwidth]{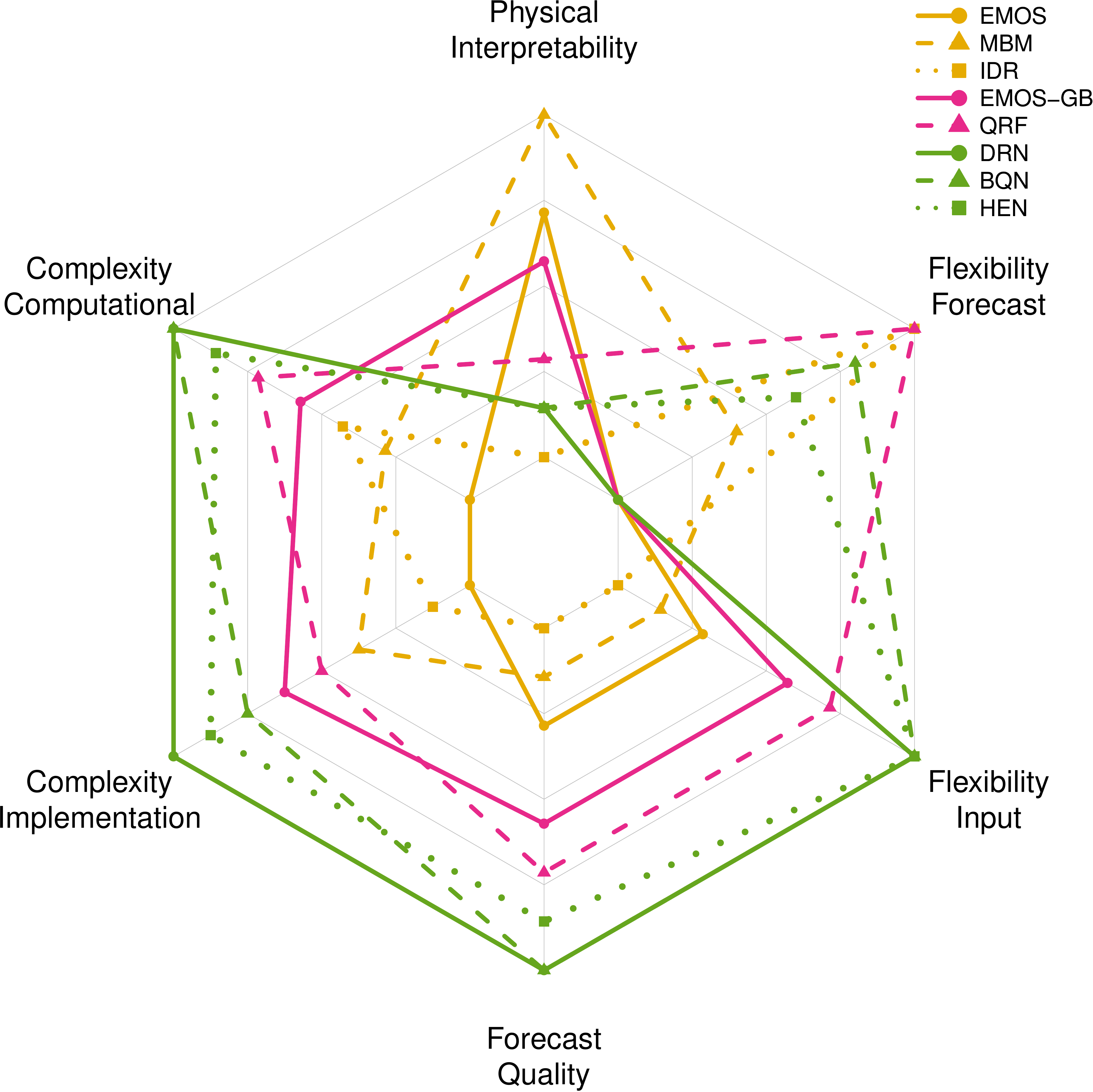}
\caption{Illustration of subjectively ranked key characteristics of the postprocessing methods presented in Section \ref{sec:methods} in form of a radar chart. In each displayed dimension, entries closer to the center indicate lower degrees (for example of forecast quality). Flexibility here refers to the flexibility of the obtained forecast distribution, or the flexibility in terms of inputs that can be incorporated into the model. The component of model complexity is divided into the computational requirements in terms of data and computing resources, and the complexity of the model implementation in terms of available software and the required choices regarding model architecture and tuning parameters. \label{fig:methods_spider_web}}
\end{center}

\bigbreak
\bigbreak

\captionof{table}{Overview of the training process and predictors used in the different postprocessing methods. The abbreviation 'Stats.' refers to summary statistics derived from the wind gust ensemble. \label{tbl:methods_training}}	
\begin{center} 
\small 
 \begin{tabular}{c|ccc|cc|cc|cc} 
\toprule
  & \multicolumn{3}{c|}{Training process} & \multicolumn{2}{c|}{Wind gust} & \multicolumn{2}{c|}{Ens.\ of other} & \multicolumn{2}{c}{Other}  \\
  & \multicolumn{3}{c|}{} & \multicolumn{2}{c|}{ensemble} & \multicolumn{2}{c|}{met.\ variables} & \multicolumn{2}{c}{predictors}  \\[1em]
Method & Estimat. & Local & Season. & Stats. & Members & Mean & Std.\,dev. & Temp. & Spatial  \\
 \hline
EMOS & CRPS & \checkmark & \checkmark & \checkmark & - & - & -  & - & - \\
MBM & CRPS & \checkmark & \checkmark & \checkmark & \checkmark & - & - & - & - \\
IDR & CRPS & \checkmark & - & - & \checkmark & -  & - & - & - \\
 \hline
EMOS-GB & MLE & \checkmark & - & \checkmark & - & \checkmark & \checkmark & \checkmark & - \\
QRF & Custom & \checkmark & - & \checkmark & - & \checkmark & - & \checkmark & - \\
 \hline
DRN & CRPS & - & - & \checkmark & -  & \checkmark & - & \checkmark & \checkmark \\
BQN & QL & - & - & - & \checkmark & \checkmark & - & \checkmark & \checkmark \\
HEN & MLE & - & - & \checkmark & - & \checkmark & - & \checkmark & \checkmark \\
\bottomrule 
\end{tabular}
\end{center}

\end{figure}

We close this section with a short discussion and overview of the general characteristics of the various postprocessing methods introduced above. For most practical applications, there does not exist a single best method as all approaches have advantages but also shortcomings. Based on our experiences for this particular study, Figure \ref{fig:methods_spider_web} presents a subjective overview of key characteristics of the different methods, ranging from flexibility and forecast quality to complexity and interpretability. The color scheme distinguishes the three groups of methods, and the different line and point styles indicate different characteristics of the forecast distributions, e.g.\ solid lines indicate the use of a parametric forecast distribution. Further, Table \ref{tbl:methods_training} presents a summary of the different predictors available to the different postprocessing approaches.

\section{Evaluation} \label{sec:eval}

In this section, we evaluate the predictive performance of the postprocessing methods based on the test period that consists of all data from 2016. Since we considered forecasts from one initialization time only, systematic changes over the lead time are closely related to the diurnal cycle. For an introduction to the evaluation methods and the underlying theory, see Appendix~\ref{sec:appendix_eval}. 

\subsection{Predictive performance of the COSMO-DE-EPS and a climatological baseline} \label{sec:eval_ens}

\begin{figure}
\begin{center}
\includegraphics[width=\textwidth]{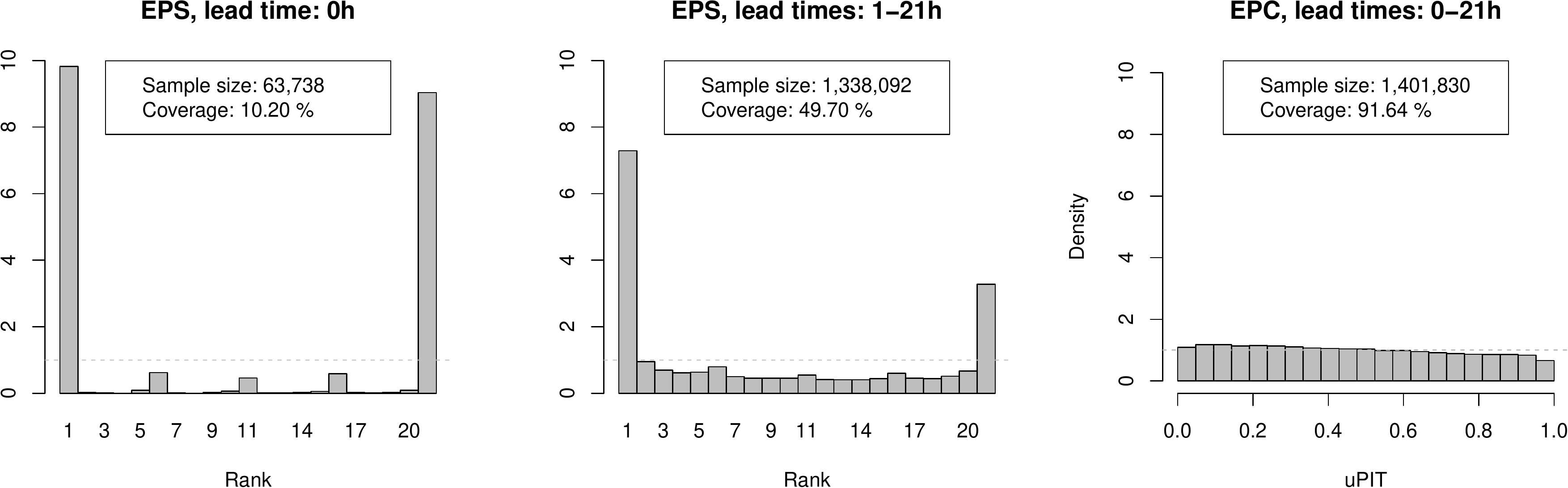}
\caption{Verification rank histograms of 0 h- and 1--21 h-forecasts of the COSMO-DE-EPS and uPIT histogram of the EPC forecasts over all lead times for all stations (left to right). Coverage refers to a prediction interval with a nominal coverage corresponding to a 20-member ensemble (ca.\ 90.48\%). \label{fig:eval_eps_epc_histograms}}
\end{center}
\end{figure}

The predictive performance of the EPS coincides with findings of extant previous studies on statistical postprocessing of ensemble forecasts in that the ensemble predictions are biased and strongly underdispersed, see Figure \ref{fig:eval_eps_epc_histograms} for the corresponding verification rank histograms.

\begin{figure}
\begin{center}
\includegraphics[width=\textwidth]{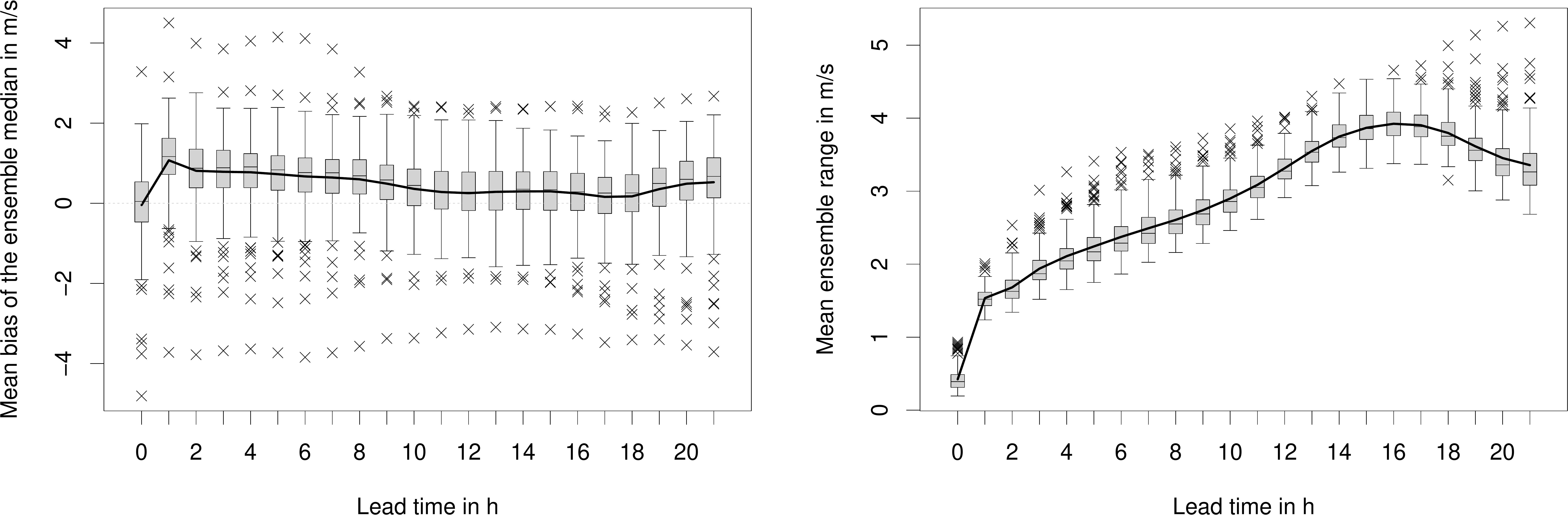}
\caption{Boxplots of the station-wise mean bias of the ensemble median (left) and the mean ensemble range (right) of the EPS as functions of the lead time. The black line indicates the average over all samples. 
\label{fig:eval_eps_bias_range}}
\end{center}\end{figure}

We here highlight two peculiarities of the EPS. The first is the so-called spin-up effect \citep[see, e.g.,][]{Kleczek2014}, which refers to the time the numerical model requires to adapt to the initial and boundary conditions and to produce structures consistent with the model physics. 
This effect can be seen not only in the ensemble range and the bias of the ensemble median prediction displayed in Figure \ref{fig:eval_eps_bias_range}, where a sudden jump at the 1 h-forecasts is observed, but also in the verification rank histograms, since there is a clear lack of ensemble spread in the 0 h-forecasts within each of the four sub-ensembles and only a small spread between them.

The temporal development of the bias and ensemble range shown in Figure \ref{fig:eval_eps_bias_range} indicates another meteorological effect, the evening transition of the planetary boundary layer. When the sun sets, the surface and low-level air that have been heated over the course of the day cool down and thermally driven turbulence ceases. This sometimes quite abrupt transition to calmer, more stable conditions strongly affects the near-surface wind fields subject to the local conditions \citep[see, e.g.,][]{Mahrt2017}. For lead times up to 18 hours (corresponding to 19:00 (20:00) local time in winter (summer)), the ensemble range increases together with an improvement in calibration. However, at the transition in the evening, the overall bias increases and the calibration becomes worse for most stations indicating increasing systematic errors.
This could be related, for example, to the misrepresentation of the inertia of large eddies in the model or errors in radiative transfer at low sun angles.

In addition to the raw ensemble predictions, we further consider a climatological reference forecast as a benchmark method. The extended probabilistic climatology \citep[EPC;][]{Vogel2018} is an ensemble based on past observations considering only forecasts at the same time of the year. We create a separate climatology for each station and hour of the day that consists of past observations from the previous, current and following month around the date of interest. The observational data base ranges back to 2001, thus EPC is built on a database of 15 years. Not surprisingly, EPC is well calibrated (see Figure \ref{fig:eval_eps_epc_histograms}). However, it shows a minor positive bias, which is likely due to the generally lower level of wind gusts observed in 2016 compared to the years on which EPC is based. 
Additional illustrations are provided in the supplemental material.

\subsection{Comparison of the postprocessing methods} \label{sec:eval_pred}

\begin{figure}
\begin{center}
\includegraphics[width=\textwidth]{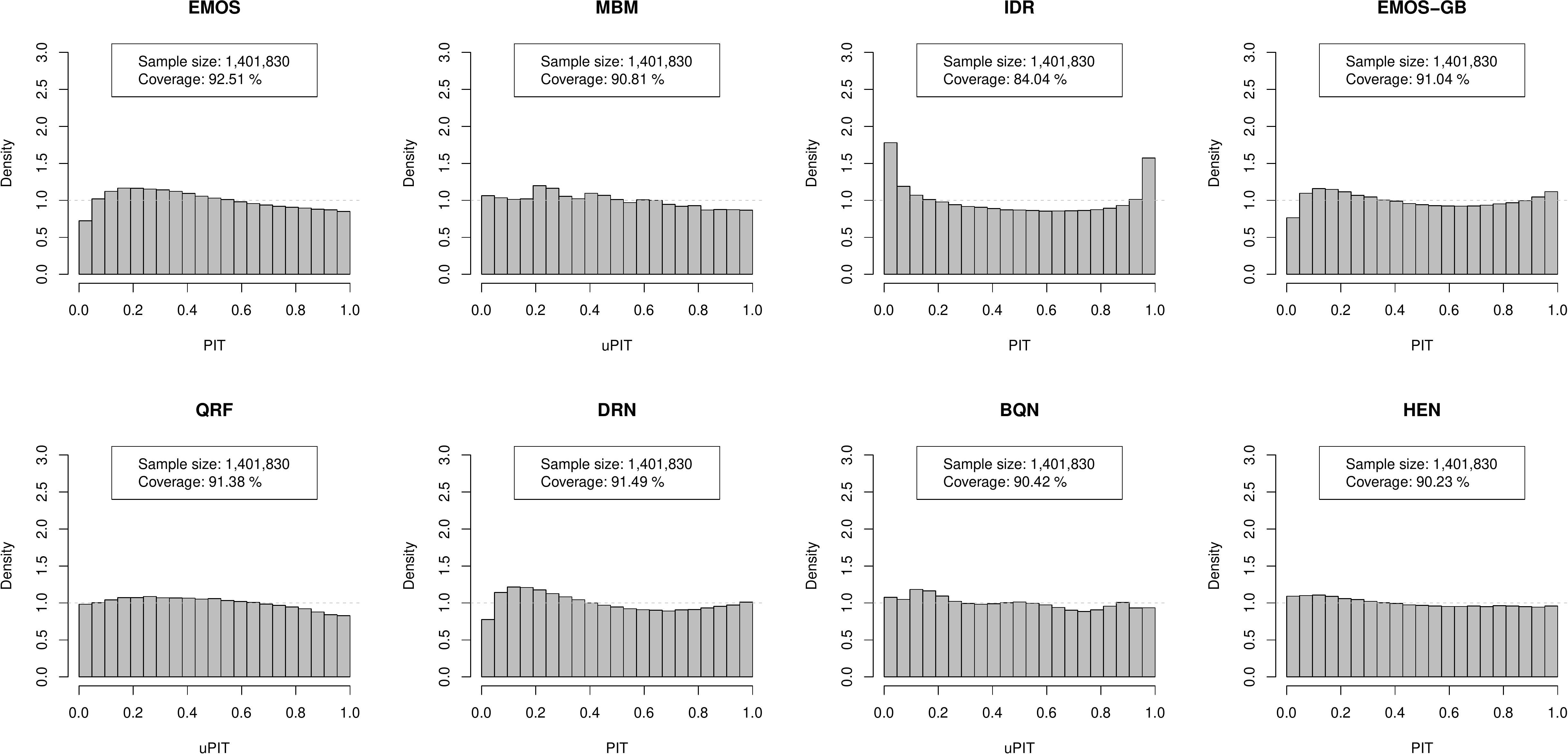}
\caption{PIT histograms of all postprocessing methods, aggregated over all lead times and stations. Coverage refers to the empirical coverage of a prediction interval with a nominal coverage corresponding to a 20-member ensemble (ca.\ 90.48\%). \label{fig:eval_histograms}}
\end{center}
\end{figure}

Figure \ref{fig:eval_histograms} shows PIT histograms for all postprocessing methods. All approaches substantially improve the calibration compared to the raw ensemble predictions and yield well calibrated forecasts, except for IDR which results in underdispersed predictions. The PIT histograms of the  parametric methods based on a truncated logistic distribution (EMOS, EMOS-GB and DRN) all exhibit similar minor deviations from uniformity caused by a lower tail that is too heavy. 
The semi- and non-parametric methods MBM, QRF, BQN and HEN are all slightly skewed to the left, in line with the histogram of EPC. Further, we observe a minor overdispersion for the QRF forecasts.

Table \ref{tbl:eval_overall_scores} summarizes the values of proper scoring rules and other evaluation metrics to compare the overall predictive performance of all methods. While the ensemble predictions outperform the climatological benchmark method, all postprocessing approaches lead to substantial improvements. Among the different postprocessing methods, the three groups of approaches introduced in Section \ref{sec:methods} show systematic differences in their overall performance. In terms of the CRPS, the basic methods already improve the ensemble by around 26--29\%. Incorporating additional predictors via the machine learning methods further increases the skill, and the NN-based approaches, in particular DRN and BQN perform best. The mean absolute error (MAE) and root mean squared error (RMSE) lead to analogous rankings, and all methods clearly reduce the bias of the EPS. Among the well-calibrated postprocessing methods, the NN-based methods yield the sharpest forecast distributions, followed by the machine learning approaches and the basic methods. 
Thus, we conclude that the gain in predictive performance is mainly based on an increase in sharpness.
Overall, BQN results not only in the best coverage, but also in the sharpest prediction intervals.

\begin{table}[t]
\caption{Evaluation metrics for EPC, COSMO-DE-EPS and all postprocessing methods averaged over all lead times and stations. PI length and coverage refer to a prediction interval with a nominal coverage corresponding to a 20-member ensemble (ca.\ 90.48\%). The best methods are indicated in bold. \label{tbl:eval_overall_scores}}	
\begin{tabular}{c@{\hskip 0.7cm}c@{\hskip 0.7cm}c@{\hskip 0.7cm}c@{\hskip 0.7cm}c@{\hskip 0.7cm}c@{\hskip 0.7cm}c@{\hskip 0.7cm}c} 
\toprule 
Method & CRPS & MAE & RMSE & Bias & PI length & Coverage & Runtime \\ 
\midrule 
EPC & 1.72 & 2.44 & 3.26 & -0.13 & 10.73 & 91.64 \% & - \\ 
EPS & 1.33 & 1.63 & 2.16 & 0.47 & 2.85 & 47.91 \% & - \\ 
\midrule 
EMOS & 0.95 & 1.32 & 1.80 & 0.05 & 5.94 & 92.51 \% & \textbf{19 min} \\ 
MBM & 0.97 & 1.34 & 1.80 & 0.04 & 6.10 & 90.81 \% & 51242 min \\ 
IDR & 0.98 & 1.36 & 1.84 & 0.01 & \textbf{4.72} & 84.04 \% & 8100 min \\ 
\midrule 
EMOS-GB & 0.88 & 1.23 & 1.69 & -0.06 & 5.24 & 91.04 \% & 510 min \\ 
QRF & 0.87 & 1.22 & 1.66 & -0.03 & 5.41 & 91.38 \% & 282 min \\ 
\midrule 
DRN & \textbf{0.84} & 1.18 & 1.61 & 0.03 & 5.05 & 91.49 \% & 399 min \\ 
BQN & 0.84 & \textbf{1.18} & \textbf{1.61} & \textbf{0.00} & 4.94 & \textbf{90.42 \%} & 387 min \\ 
HEN & 0.86 & 1.21 & 1.64 & -0.04 & 5.07 & 90.23 \% & 321 min \\ 
\bottomrule 
\end{tabular}
\end{table}

We further consider the total computation time required for training the postprocessing models. However, note that a direct comparison of computation times is difficult due to the differences in terms of software packages and parallelization capabilities. Not surprisingly, the simple EMOS method was the fastest with only 19 minutes. The network-based methods were not much slower than QRF and faster than EMOS-GB, which is based on almost twice as much predictors as the other advanced methods what approximately doubled the computational costs. 
MBM here is an extreme outlier and requires a computation time of over 35 days in total, in particular due to our adaptions to the sub-ensemble structure discussed in Section \ref{sssec:methods_mbm}.

\begin{figure}[p]
\begin{center}
\includegraphics[width=\textwidth]{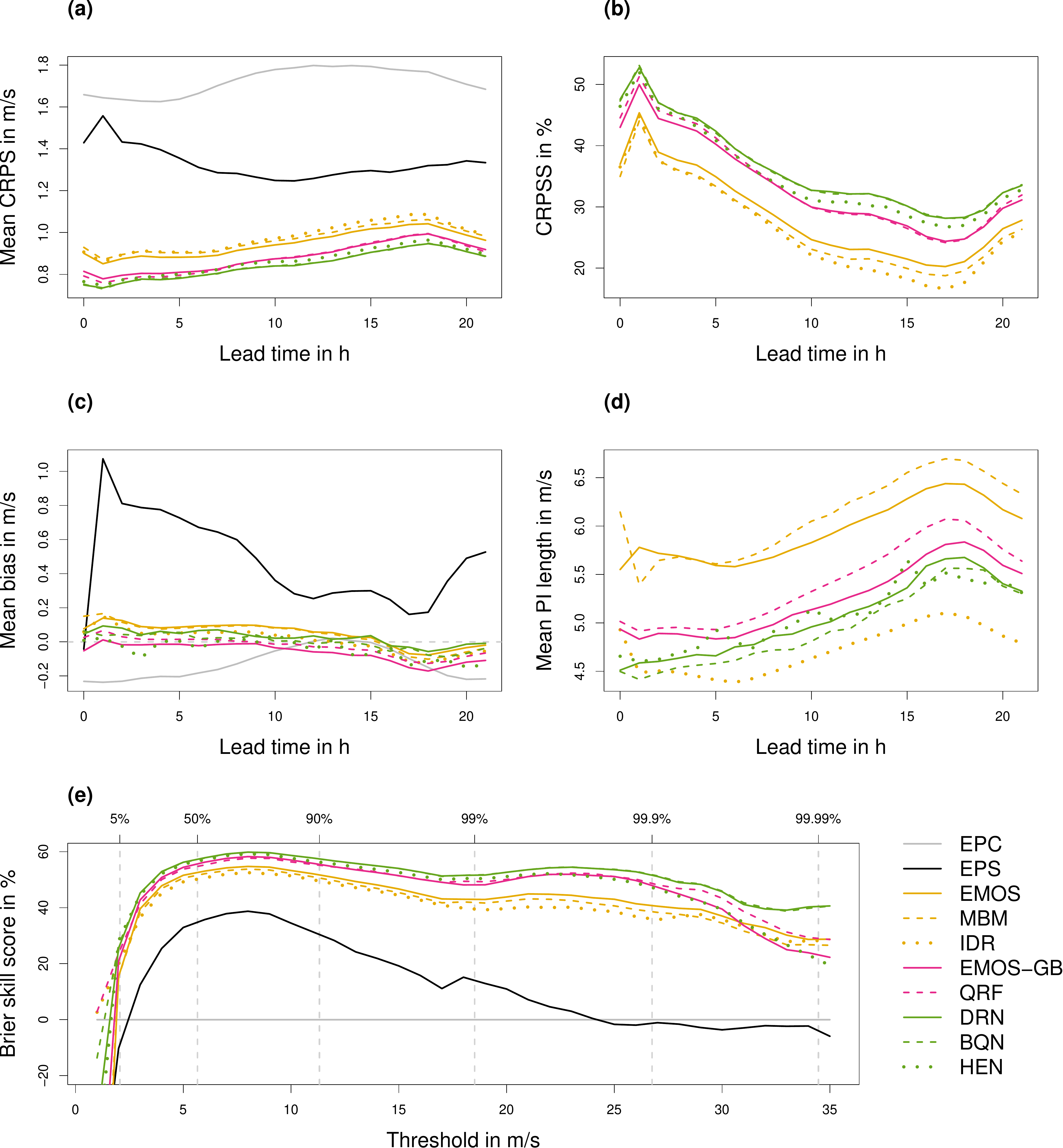}
\caption{Mean CRPS (a), CRPSS w.r.t.\ the raw ensemble predictions (b), mean bias (c) and mean prediction interval length (d) of the postprocessing methods as functions of the lead time, averaged over all stations. Panel (e) shows the Brier skill score w.r.t.\ the climatological EPC forecast as function of the threshold, averaged over all lead times and stations, where the dashed vertical lines indicate the quantiles of the observed wind gusts at levels given at the top axis.
\label{fig:eval_lead_time_panel}}
\end{center}
\end{figure}

\subsection{Lead time-specific results}

To investigate the effects of the different lead times and hours of the day on the predictive performance, Figure \ref{fig:eval_lead_time_panel} shows various evaluation metrics as function of the lead time. 
While the CRPS values and the improvements over the raw ensemble predictions (panels a and b) show some variations over the lead times, the overall rankings among the different methods and groups of approaches are consistent. In particular, the rankings of the individual postprocessing models remain relatively stable over the day.

The spin-up effect is clearly visible in that the mean bias drastically increases from the 0 h to the 1 h-forecasts of the EPS and leads to a worse CRPS despite the increase of the ensemble range (see Figure \ref{fig:eval_eps_bias_range}). 
The postprocessed forecasts, however, are able to successfully correct the biases induced in the spin-up period while benefiting from the increase in ensemble range. Hence, the CRPS becomes smaller and the skill is the largest through all lead times. 
Although we improve the MBM forecast by incorporating the sub-model structure, the adjusted ensemble forecasts are still subject to systematic deviations from calibration for lead times of 0 and 1 h.
Additional illustrations are provided in the supplemental material.

Following the first hours, a somewhat counter-intuitive trend can be observed in that the predictive performance of the EPS improves up to a lead time of 10 h. This is in particular due to improvements in terms of the spread of the ensemble over time. By contrast, the predictive performance of the climatological baseline model is affected more by the diurnal cycle since observed wind gusts and their variability tend to be higher during daytime. 
The performance of the postprocessing models is neither affected by the increased spread of the EPS, nor by larger gust observations, and slightly decreases over time until the evening transition. This is in line with wider prediction intervals that represent increasing uncertainty for longer lead times, while the mean bias and coverage are mostly unaffected.

This general trend changes with the evening transition at a lead time of around 18 h. The CRPS of the climatological reference model decreases due to a better predictability of the wind gust forecasts. By contrast, the CRPS of the ensemble increases, again driven by an increase in bias and a decrease in spread that comes with a smaller coverage. 
The numerical model thus appears to not be fully capable of capturing the relevant physical effects, and introduces systematic errors.
The bias and coverage of the postprocessing methods do not change drastically, while the prediction intervals of the postprocessing methods become smaller, which is in line with the more stable conditions at nighttime. Therefore, the CRPS of the postprocessing methods becomes better again.

To assess the forecast performance for extreme wind gust events, Figure \ref{fig:eval_lead_time_panel}e shows the mean Brier skill score w.r.t.\ the climatological EPC forecast as a function of the threshold value, averaged over all stations and lead times. For larger threshold values, the EPS rapidly looses skill and does not provide better predictions than the climatology for thresholds above 25 m/s. By contrast, all postprocessing methods retain positive skill across all considered threshold values. The predictive performance decreases for very high threshold values above 30 m/s, in particular for EMOS-GB and QRF. Note that the EPS and all postprocessing methods besides the analog-based QRF and IDR have negative skill scores for very small thresholds, but this is unlikely to be of relevance for any practical application.

\subsection{Station-specific results and statistical significance}

\begin{figure}[p]
\captionof{table}{Ratio of lead time-station combinations (in \%) where pairwise DM tests indicate statistically significant CRPS differences after applying a Benjamini-Hochberg procedure to account for multiple testing for a nominal level of $\alpha = 0.05$ of the corresponding one-sided tests. The $(i, j)$-entry in the $i$-th row and $j$-th column indicates the ratio of cases where the null hypothesis of equal predictive performance of the corresponding one-sided DM test is rejected in favor of the model in the $i$-th row when compared to the model in the $j$-th column. The remainder of the sum of $(i, j)$- and $(j, i)$-entry to 100\% is the ratio of cases where the score differences are not significant. \label{tbl:eval_significance_table}}	
\begin{center}  
	\scalebox{0.95}{ \begin{tabular}{ccccccccccc} 
\toprule
  & EPC & EPS & EMOS & MBM & IDR & EMOS-GB & QRF & DRN & BQN & HEN \\ 
  \midrule
  EPC &  & 1.7 & 0.1 & 0.2 & 0.3 & 0.0 & 0.0 & 0.0 & 0.0 & 0.0 \\ 
  EPS & 70.8 &  & 0.0 & 0.0 & 0.0 & 0.0 & 0.0 & 0.0 & 0.0 & 0.0 \\ 
  EMOS & 99.5 & 100.0 &  & 97.6 & 83.4 & 1.9 & 1.0 & 0.1 & 0.2 & 1.0 \\ 
  MBM & 99.4 & 99.9 & 0.8 &  & 51.3 & 0.1 & 0.2 & 0.0 & 0.0 & 0.2 \\ 
  IDR & 98.6 & 99.7 & 4.3 & 20.8 &  & 0.4 & 0.1 & 0.0 & 0.0 & 0.1 \\ 
  EMOS-GB & 100.0 & 100.0 & 88.8 & 96.2 & 94.9 &  & 32.9 & 3.2 & 3.4 & 11.7 \\ 
  QRF & 100.0 & 100.0 & 89.7 & 97.2 & 98.3 & 35.2 &  & 5.3 & 5.3 & 13.0 \\ 
  DRN & 100.0 & 100.0 & 98.1 & 99.2 & 99.4 & 84.4 & 81.0 &  & 46.6 & 83.4 \\ 
  BQN & 100.0 & 100.0 & 97.9 & 99.2 & 99.4 & 84.1 & 80.4 & 43.3 &  & 83.5 \\ 
  HEN & 99.9 & 100.0 & 95.1 & 97.7 & 97.7 & 67.6 & 64.2 & 8.2 & 8.0 &  \\ 
  \bottomrule
\end{tabular} }
\end{center}

\bigbreak
\bigbreak
\bigbreak

\begin{center}
\includegraphics[width=0.63\textwidth]{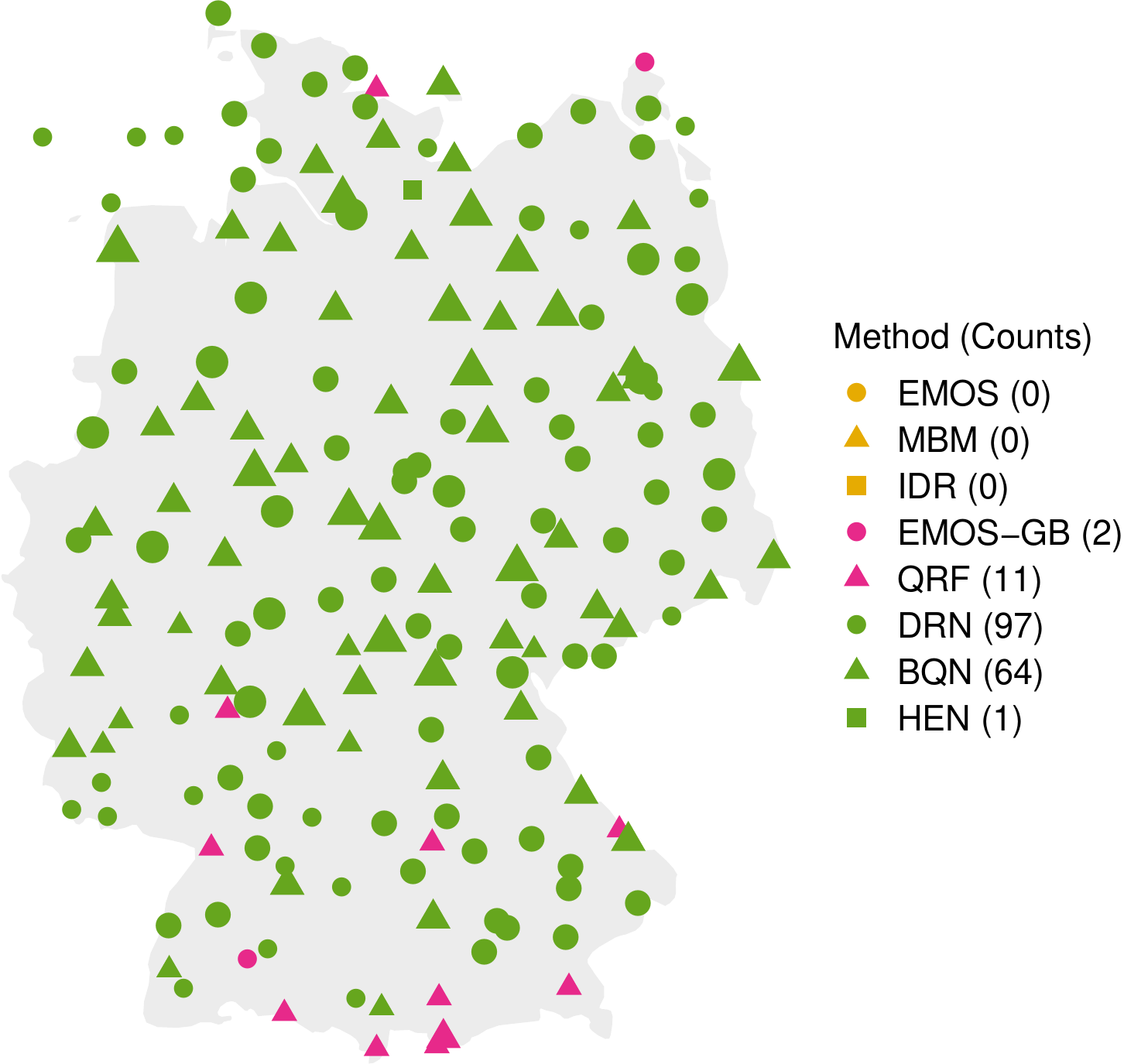}
\captionof{figure}{Best method at each station in terms of the CRPS, averaged over all lead times. 
The point sizes indicate the level of statistical significance of the observed CRPS differences compared to the methods only from the other groups of methods for all lead times. Three different point sizes are possible, with the smallest size indicating statistically significant differences for at most 90\% of the performed tests, the middle size for up to 99\% and the largest to 100\% meaning all differences are statistically significant. 
\label{fig:eval_map_best_crps}}
\end{center}

\end{figure}

We further investigate the station-specific performance of the different postprocessing models and in particular investigate whether the locally adaptive networks that are estimated jointly for all stations also outperform the locally estimated methods at the individual stations. Figure \ref{fig:eval_map_best_crps} shows a map of all observation stations indicating the station-specific best model, and demonstrates that at 162 of the 175 stations a network-based method performs best. While none of the basic methods provides the best forecasts at any station, QRF or EMOS-GB perform best at the remaining 13 stations. Most of these station are located in mountainous terrain or coastal regions that are likely subject to specific systematic errors, which might favor a location-specific modeling approach. 

Finally, we evaluate the statistical significance of the differences in the predictive performance in terms of the CRPS between the competing postprocessing methods. To that end, we perform Diebold-Mariano (DM) tests of equal predictive performance \citep{Diebold1995} for each combination of station and lead time, and apply a \citet{Benjamini1995} procedure to account for potential temporal and spatial dependencies of forecast errors in this multiple testing setting. Mathematical details are provided in Appendix \ref{sec:appendix_eval}. 

We find that the observed score differences are statistically significant for a high ratio of stations and lead times, see Table \ref{tbl:eval_significance_table}. In particular, DRN and BQN significantly outperform the basic models at more than 97\%, and even significantly outperform the machine learning methods at more than 80\% of all combinations of stations and lead times. Among the basic and machine learning methods, QRF performs best but only provides significant improvements over the NN-based methods for around 5\% of the cases.

To assess station-specific effects of the statistical significance of the score differences, the sizes of the points indicating the best models in Figure \ref{fig:eval_map_best_crps} are scaled by the degree of statistical significance of the results when compared to all models from the two other groups of methods. For example, if DRN performs best at a station, the corresponding point size is determined by the proportion of rejections of the null hypothesis of equal predictive performance at that station when comparing DRN to EMOS, MBM, IDR, EMOS-GB and QRF (but not the other NN-based models) for all lead times in a total of $5\cdot 22$ DM tests. 
Generally, if a locally estimated machine learning approach performed best at one station, the significance tends to be lower than when a network-based method performs best. The most significant differences between the groups of methods can be observed in central Germany, where most stations likely exhibit similar characteristics compared to coastal areas in northern Germany or mountainous regions in southern Germany.

\section{Feature importance} \label{ssec:eval_features}

The results presented in the previous section demonstrate that the use of additional features improves the predictive performance by a large margin. Here, we assess the effects of the different inputs on the model performance to gain insight into the importance of meteorological variables and better understand what the models have learned. Many techniques have been introduced in order to better interpret machine learning methods, in particular NNs \citep{McGovern2019}, and we will focus on distinct approaches tailored to the individual machine learning methods at hand. 

\subsection{EMOS-GB and QRF}

Since the second group of methods relies on locally estimated, separate models for each station, the importance of specific predictors will often vary across different locations and thus make an overall interpretation of the model predictions more involved.
 
In the case of EMOS-GB, we treat the location and scale parameters separately and consider a feature to be more important the larger the absolute value of the estimated coefficient value is. In the interest of brevity, we here discuss some general properties only, and refer to the supplemental material for  detailed results. Overall, the interpretation is in particular challenging due to a large variation across stations, in particular during the spin-up period.
In general, the mean value of the wind gust predictions is selected as the most important predictor for the location parameter, followed by other wind-related predictors and the temporal information about the day of the year. For the scale parameter, the standard deviation of the ensemble predictions of wind gust is selected as the most important predictor, followed by the ensemble mean. Other meteorological predictors tend to only contribute relevant information for specific combinations of lead times and stations, parts of which might be physically inconsistent coefficient estimates due to random effects in the corresponding datasets. Selected examples are presented in the supplemental material.

For QRF, we utilize an out-of-bag estimate of the feature importance based on the training set \citep{Breiman2001}. The procedure is similar to what we apply for the NN-based models below, but uses a different evaluation metric directly related to the algorithm for constructing individual decision trees, see \citet{Wright2017ranger} for details. Figure \ref{fig:fi_qrf} shows the feature importance for some selected predictor variables as a function of the lead time; additional results are available in the supplemental material. Interestingly, the ten most important predictors (two of which are included in Figure \ref{fig:fi_qrf}) are variables that directly relate to different characteristics of wind speed predictions from the ensemble. This can be explained by the specific structure of random forest models. Since these predictor variables are highly correlated, they are likely to serve as replacements if other ones are not available in the random selection of potential candidate variables for individual splitting decisions. 
The standard deviation of the wind gust ensemble is only of minor importance during the spin-up period.
Besides the wind-related predictors from the EPS, the day of the year, the net short wave radiation flux prediction as well as the relative humidity prediction at 1,000 hPa are selected as important predictors, particularly for longer lead times corresponding to later times of the day and potentially again indicating an effect of the evening transition.
In particular the short wave radiation flux indicates the sensitivity of the wind around sunset to the maintenance of turbulence by surface heating, an effect not seen in the morning when the boundary layer grows more gradually.

\subsection{Neural network-based methods}

To investigate the feature importance for NNs, we follow \cite{Rasp2018} and use a permutation-based measure that is given by the decrease in terms of the CRPS in the test set when randomly permuting a single input feature, using the mean CRPS of the model based on unpermuted input features as reference. In order to eliminate the effect of the dependence of the forecast performance on the lead times, we calculate the relative permutation importance. Mathematical details are provided in Appendix \ref{sec:appendix_eval}.

\begin{figure}[p]
\begin{center}
\includegraphics[width=0.9\textwidth]{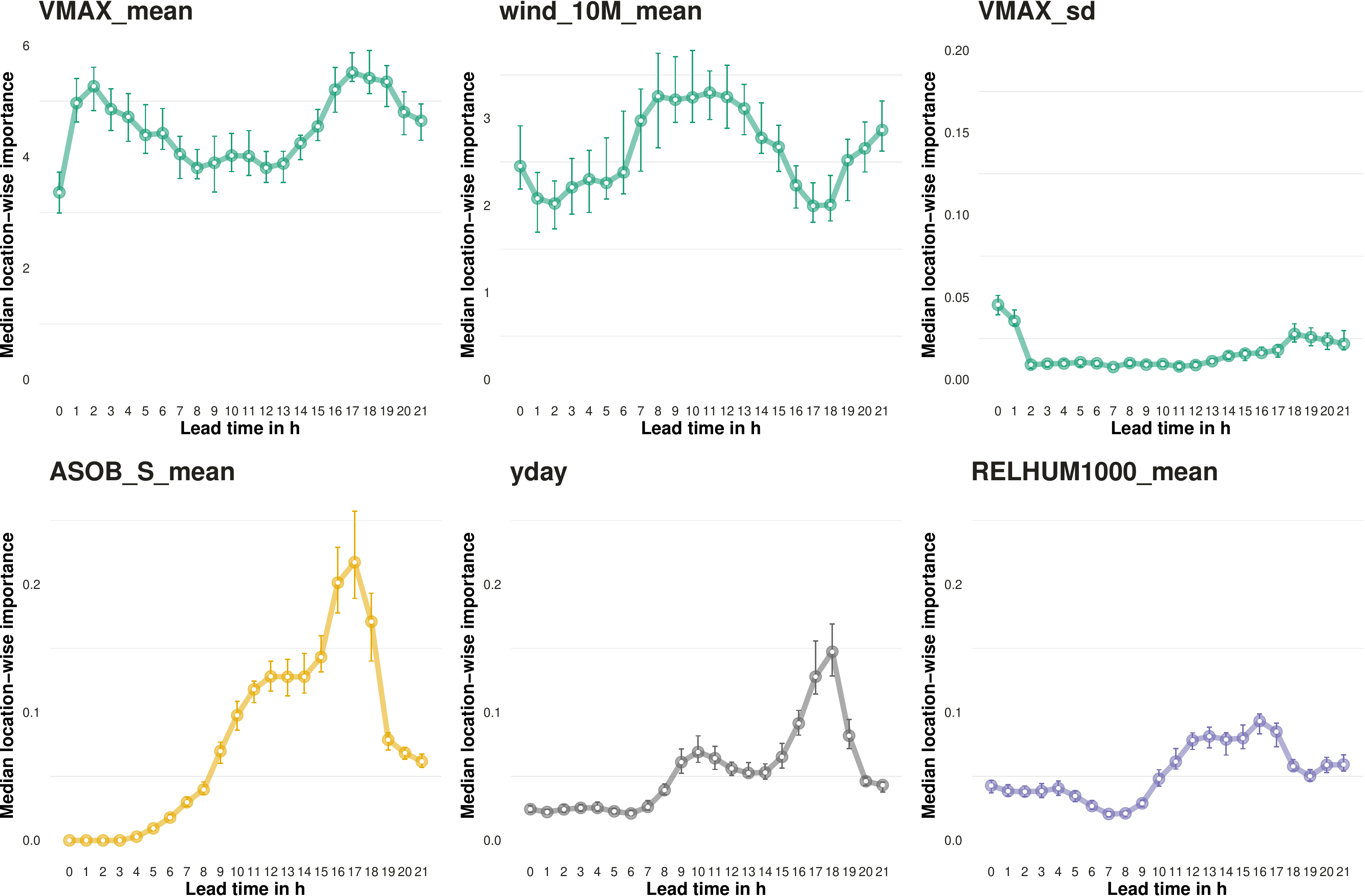}
\caption{Median of station-wise feature importance for selected predictors (see Table \ref{tbl:data_predictors}) of the QRF model as functions of the lead time. The error bars indicate a bootstrapped 95\%-confidence interval of the median. Note the different scale of the vertical axes. 
\label{fig:fi_qrf}}

\bigbreak 

\includegraphics[width=0.9\textwidth]{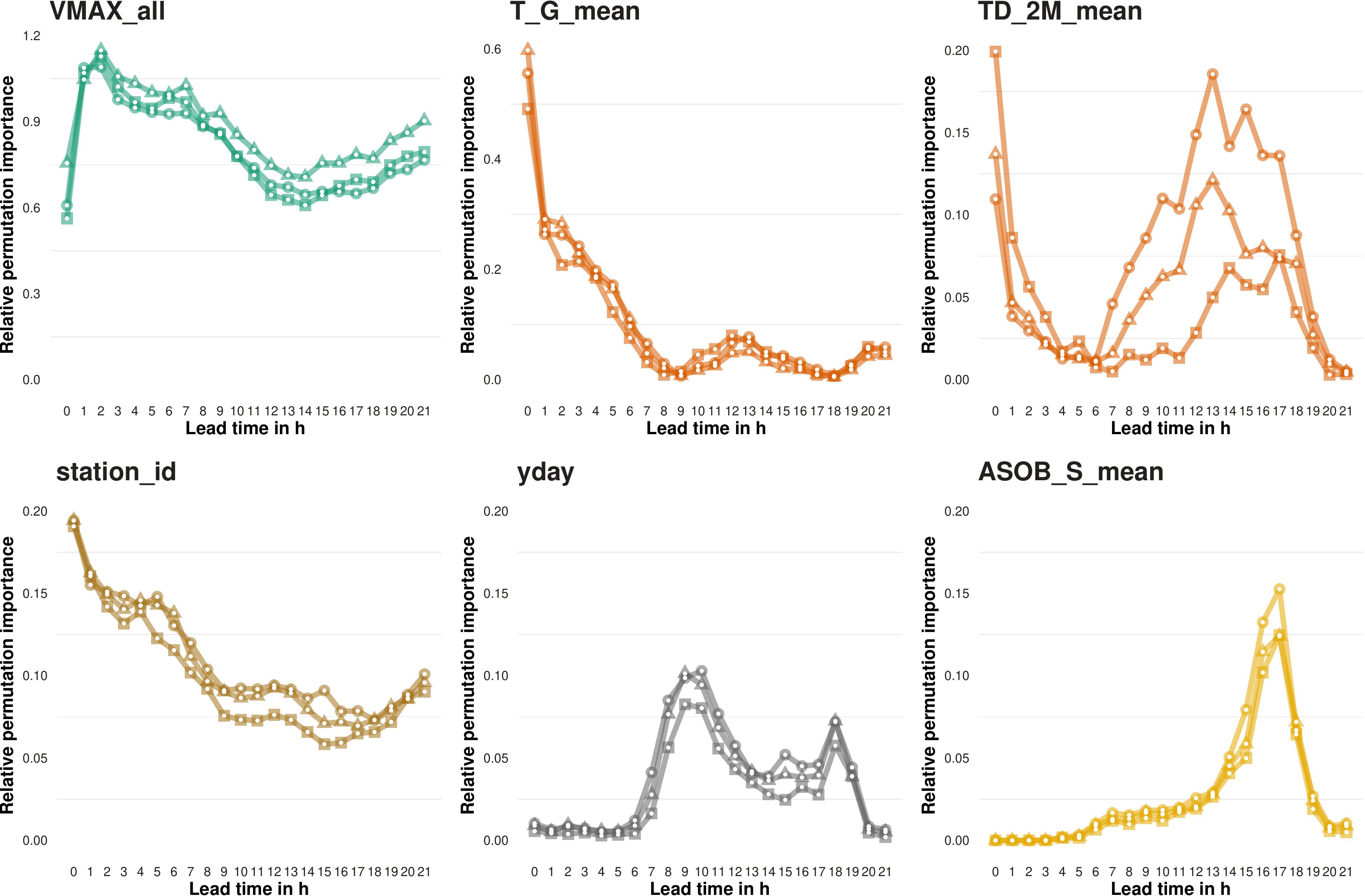}
\caption{Relative permutation importance of selected predictors (see Table \ref{tbl:data_predictors}) for the three NN-based models dependent on the lead time. Note the different scale of the vertical axes. 
The abbreviation \code{VMAX\_all} refers to the multi-pass permutation of the features derived from the wind gust ensemble.
Different symbols indicate the three model variants ($\circ$: DRN, $\triangle$: BQN, $\square$: HEN). \label{fig:eval_fi_nn_features}}
\end{center}
\end{figure}

Figure \ref{fig:eval_fi_nn_features} shows the relative permutation importance for selected input features and the three NN-based postprocessing methods; additional results are provided in the supplemental material. There are only minor variations across the three NN approaches, with the wind gust ensemble forecasts providing the most important source of information. To ensure comparability of the three model variants, we here jointly permute the corresponding features of the ensemble predictions of wind gust (mean and standard deviation for DRN and HEN, and the sorted ensemble forecast for BQN). Further results for BQN available in the supplemental material indicate that among the sorted ensemble members, the minimum and maximum value are the most important member predictions, followed by the ones indicating transitions between the groups of sub-ensembles.
Again, we find that the standard deviation of the wind gust ensemble forecasts is of no importance for DRN and HEN (not shown).

In addition to the wind gust ensemble predictions, the spatial features form the second most important group of predictors. The station ID (via embedding), altitude and station-wise bias are the most relevant spatial features and have a diurnal trend that resembles the mean bias of the EPS forecasts, indicating that the spatial information becomes more relevant when the bias in the EPS is larger. Further, the day of the year and the net short wave radiation flux at the surface provide relevant information that can be connected to the previously discussed evening transition as well as the diurnal cycle. 
Several temperature variables, in particular temperature at the ground-level and lower levels of the atmosphere, constitute important predictors for different times of day, e.g.\ the ground-level temperature is important for the first few lead times during early morning.

\section{Discussion and conclusion} \label{sec:conclusions}

We have conducted a comprehensive and systematic review and comparison of statistical and machine learning methods for postprocessing ensemble forecasts of wind gusts. The postprocessing methods can be divided into three groups of approaches of increasing complexity ranging from basic methods using only ensemble forecasts of wind gusts as predictors to machine learning methods and NN-based approaches. While all yield calibrated forecasts and are able to correct the systematic errors of the raw ensemble predictions, incorporating information from additional meteorological predictor variables leads to significant improvements in forecast skill. 
In particular, postprocessing methods based on NNs jointly estimating a single, locally adaptive model at all stations provide the best forecasts and significantly outperform benchmark methods from machine learning. 
The flexibility of the NN-based methods is exemplified by the proposed modular framework that nests three variants with different probabilistic forecast types obtained as output. While DRN forecasts are given by a parametric distribution, BQN performs a quantile function regression based on Bernstein polynomials and HEN approximates the predictive density via histograms. The analysis of feature importance for the advanced methods in Section \ref{ssec:eval_features} illustrates that the machine learning techniques, in particular the NN approaches, learn physically consistent relations.
Overall, our results underpin the conjecture of \citet{Rasp2018} who argue that NN-based methods will provide valuable tools for many areas of statistical postprocessing and forecasting.

From an operational point of view, one major shortcoming of the postprocessing methods other than MBM is that they do not preserve spatial, temporal or multi-variable dependencies in the ensemble forecasts, which is one of the main challenges in postprocessing \citep{Vannitsem2021}. Several studies have investigated approaches that are able to reconstruct the correlation structure \citep[see e.g.][]{Schefzik2013,Lerch2020}, however these techniques require additional steps and therefore increase the complexity of the postprocessing framework. While all methods considered here can form basic building blocks for such multivariate approaches, an interesting avenue for future work is to combine the MBM approach with NNs, which might allow to efficiently incorporate information from additional predictor variables while preserving the physical characteristics. 

A related limitation of the postprocessing methods considered here is that they are not seamless in space and time as they rely on separate models for each lead time, and even each station in case of the basic approaches, as well as EMOS-GB and QRF. In practice, this may lead to physically inconsistent jumps in the forecast trajectories. To address this challenge, \cite{Keller2021} propose a global EMOS variant that is able to incorporate predictions from multiple NWP models in addition to spatial and temporal information. For the NN-based framework for postprocessing considered here, an alternative approach to obtain a joint model across all lead times would be to embed the temporal information in a similar manner as the spatial information.

The postprocessing methods based on NNs provide a starting point for flexible extensions in future research. In particular, the rapid developments in the machine learning literature offer unprecedented ways to incorporate new sources of information into postprocessing models, including spatial information via convolutional neural networks \citep{Scheuerer2020, Veldkamp2021}, or temporal information via recurrent neural networks \citep{Gasthaus2019}. A particular challenge for weather prediction is given by the need to better incorporate physical information and constraints into the forecasting models. Physical information about large-scale weather conditions, or weather regimes, forms a particularly relevant example in the context of postprocessing \citep{Rodwell2018}, with recent studies demonstrating benefits of regime-dependent approaches \citep{Allen2020,Allen2021}. For wind gusts in European winter storms, \cite{Pantillon2018} found that a simple EMOS approach may substantially deteriorate forecast performance of the raw ensemble predictions during specific meteorological conditions. An important step for future work will be to investigate whether similar effects occur for the more complex approaches here, and to devise dynamical feature-based postprocessing methods that are better able to incorporate relevant domain knowledge by tailoring the model structure and estimation process.

\section*{Acknowledgments}
The research leading to these results has been done within the project C5 ``Dynamical feature-based ensemble postprocessing of wind gusts within European winter storms'' of the Transregional Collaborative Research Center SFB/TRR 165 ``Waves to Weather'' funded by the German Research Foundation (DFG). Sebastian Lerch gratefully acknowledges support by the Vector Stiftung through the Young Investigator Group ``Artificial Intelligence for Probabilistic Weather Forecasting''. We thank Reinhold Hess and Sebastian Trepte for providing the forecast and observation data, and Robert Redl for assistance in data handling. We further thank Lea Eisenstein, Tilmann Gneiting, Alexander Jordan and Peter Knippertz for helpful comments and discussions, as well as John Bremnes for providing code for the implementation of BQN. 

\bibliographystyle{myims2}
\bibliography{paper_pp_gusts_mod}

\newpage

\appendix

\section{Forecast evaluation}\label{sec:appendix_eval}

We here provide a summary of the methods used for forecast evaluation. In the following, we will refer to a probabilistic forecast by $F$, to the random variable of the observation by $Y$ and to a realization of $Y$ by $y$, i.e.\ an observed wind gust.

\subsubsection*{Calibration and sharpness}

We will evaluate the ensemble and the postprocessed forecasts guided by the principle that a probabilistic forecast should aim to maximize sharpness, subject to calibration \citep{Gneiting2007probabilistic}. A probabilistic forecast is said to be \textit{calibrated} if it statistically consistent with the observations, \textit{sharpness} refers to the concentration of the forecast distribution and is a property of the forecast alone. 

Qualitatively, the calibration of a probabilistic forecast can be assessed via histograms of the joint distribution of forecast and observation based on the idea that the \textit{probability integral transform} (PIT) $F(Y)$ is standard uniformly distributed if the forecast is calibrated. When only a set of quantiles or the quantile function is given, we can calculate the \textit{unified PIT} (uPIT), a generalized version of the PIT \citep{Vogel2018}. In case of ensemble forecasts, plots of the rank of the observations in the ordered ensemble provide a similar illustration. Systematic deviations from uniformity indicate specific deficiencies such as biases or dispersion errors, e.g.\ in case of a U-shaped histogram we refer to an \textit{underdispersed} forecast and in case of a hump-shaped histogram to an \textit{overdispersed} forecast. Calibration can also be assessed quantitatively via the empirical coverage of a prediction interval (PI). Given a $(1 - \alpha)$\%-PI, we can calculate the empirical coverage as the ratio of observations that fall within the corresponding PI. If the forecast is calibrated, $(1 - \alpha)$\% of the observations are expected to fall within the range of the PI. In case of an ensemble of size $m$, the ensemble range is considered a $(m-1)/(m+1)$\%-PI. Sharpness can be assessed based on the length of a PI. The shorter a PI is, the more concentrated the forecast distribution and thus the sharper the forecast.

\subsubsection*{Proper scoring rules}

Calibration and sharpness can be assessed simultaneously with proper scoring rules, which allow for a quantitative comparison of probabilistic forecasts \citep{Gneiting2007scoring}. A (negatively oriented) \textit{scoring rule} $S$ is a function that assigns a penalty to a forecast-observation tuple $(F, y)$. It is said to be \textit{proper} relative to a class of probabilistic forecasts $\mathcal{F}$ if the expected score is minimized by the underlying distribution of the observation, i.e.\
\begin{eqnarray*}
	\mathbb{E}_{G} S \left( G, Y \right) \leq \mathbb{E}_{G} S \left( F, Y \right), \quad \forall\, F,G \in \mathcal{F}.
\end{eqnarray*}
It is said to be \textit{strictly proper} if equality holds if and only if $F = G$.

One of the most prominent strictly proper scoring rules in atmospheric sciences is the \textit{continuous ranked probability score} \citep[CRPS;][]{Matheson1976}
\begin{eqnarray*}
	\text{CRPS} (F, y) = \int_{-\infty}^{\infty} \left( F(z) - \mathbbm{1} \lbrace y \leq z \rbrace \right)^2 dz,
\end{eqnarray*}
for forecast distributions $F$ with finite first moment. The CRPS is given in the same unit as the observation and generalizes to the absolute error in case of a deterministic forecast. The integral can be calculated analytically for a wide range of forecast distributions, e.g.\ for a truncated logistic distribution \citep[see, e.g.,][]{Jordan2019scoringrules}. 
In the supplemental material, we derive the CRPS of a piecewise uniform distribution based on \citet{Jordan2016}. Another popular strictly proper scoring rule is the \textit{log-score} \citep[LS;][]{Good1952} or \textit{ignorance score}
\begin{eqnarray*}
	\text{LS} (F, y) = - \log \left( f(y) \right),
\end{eqnarray*}
where $F$ is a probabilistic forecast with density $f$. 

In practical situations, a probabilistic forecast might be reduced to a single value via a statistical functional such as the mean, median or a quantile. For such cases, so-called consistent scoring functions provide useful techniques for forecast evaluation and induce corresponding proper scoring rules, see \citet{Gneiting2011points} for details. In particular, we use the \textit{quantile loss} at level $\tau \in (0,1)$ to evaluate a quantile forecast $Q_F^\tau$ at level $\tau$
\begin{eqnarray*}
	\rho_\tau (F, y) = \left(Q_F^\tau - y\right) \left( \mathbbm{1} \lbrace Q_F^\tau \geq y \rbrace - \tau \right),
\end{eqnarray*}
and the squared error for the mean forecast $\text{mean}(F)$
\begin{eqnarray*}
	\text{SE} (F, y) = \left( \text{mean}(F) - y \right)^2.
\end{eqnarray*}
Apart from proper scoring rules, we further employ the forecast error
\begin{eqnarray*}
	\text{FE} (F, y) = \text{median}(F) - y,
\end{eqnarray*}
in combination with the median forecast $\text{median}(F)$ to check the bias of the forecast distribution.

\subsubsection*{Optimum score estimation}

Strictly proper scoring rules are not only used for the comparison of probabilistic forecasts but also for parameter estimation, which is then referred to as optimum score estimation \citep{Gneiting2007scoring}. Let $F (\pmb{x}; \theta)$ be a parametric forecast distribution dependent on the predictor variables $\pmb{x}$ and parameter (vector) $\theta \in \Theta$, where $\Theta$ is the parameter space. We can estimate the optimal parameter (vector) $\theta$ by minimizing the mean score of a strictly proper scoring rule $S$, i.e.\
\begin{eqnarray*}
	\hat{\theta} = \arg\min_{\theta \in \Theta} \dfrac{1}{n} \sum_{i = 1}^n S (F (\pmb{x}_i; \theta), y_i),
\end{eqnarray*}
where $(\pmb{x}_i, y_i)$, $i = 1, \dots, n$, denotes a training set of size $n$. Note that minimizing the LS is equivalent to maximum likelihood estimation.

\subsubsection*{Statistical tests of equal predictive performance}

Given a strictly proper scoring rule $S$, forecasting methods are ranked by their average score,
\begin{eqnarray*}
	\overline{S}^F_n = \dfrac{1}{n} \sum_{i = 1}^n S (F_i, y_i),
\end{eqnarray*}
for a set of $n$ forecasts with corresponding observations. 

Comparing two forecasting methods $F$ and $G$, we can perform a statistical test of equal predictive performance via the \textit{Diebold-Mariano test} \citep{Diebold1995}. If the forecast cases are independent, the corresponding test statistic is given by
\begin{eqnarray*}
	t_n = \sqrt{n} \dfrac{\overline{S}^F_n - \overline{S}^Q_n}{\hat{\sigma}_n},
\end{eqnarray*}
where
\begin{eqnarray*}
	\hat{\sigma}_n = \dfrac{1}{n} \sum_{i = 1}^n\left( S (F_i, y_i) - S (G_i, y_i) \right)^2.
\end{eqnarray*}
Subject to weak regularity conditions, the test statistic is asymptotically standard normal under the null hypothesis of equal predictive performance, i.e.\ $\overline{S}^F_n - \overline{S}^Q_n = 0$. The null hypothesis is rejected for large (absolute) values of $t_n$, where method $F$ is preferred if the test statistic is negative, $G$ if it is positive. In our case, we perform tests of statistical significance based on the CRPS for each station and lead time separately.

Following suggestions of \cite{Wilks2016}, we apply a \textit{Benjamini-Hochberg procedure} \citep{Benjamini1995} that allows to account for multiple testing and to control the false discovery rate, which we choose to be $\alpha = 0.05$. Given the ordered $p$-values $p_{(1)}, \dots, p_{(M)}$ of $M$ hypothesis tests, a threshold $p$-value is determined via
\begin{eqnarray*}
	p^* = p_{(i^*)}, \quad \text{where} \quad i^* = \min \lbrace i = 1, \dots, M: p_{(i)} \leq \alpha \cdot i/M \rbrace.
\end{eqnarray*}
This threshold $p^*$ is then used to decide whether the null hypothesises of the individual tests are rejected. 
For Table \ref{tbl:eval_significance_table}, we applied the Benjamini-Hochberg correction for each pair of methods separately considering tests for each combination of location and lead time. For Figure \ref{fig:eval_map_best_crps}, we applied the correction for each location separately considering the tests comparing the best method in terms of the CRPS with the methods from the other groups for all lead times.

\subsubsection*{Permutation importance}

To introduce the notion of \textit{permutation importance} \citep{Rasp2018,McGovern2019}, we use $\xi$ to denote the $i$-th predictor ($i = 1, \dots, p$), $F(\pmb{x}_{\cdot j})$ ($j = 1, \dots, n$) to denote the probabilistic forecast generated by postprocessing method based on the $j$-th sample of a test set of size $n$, and $\pi$ to denote a random permutation of the set $\lbrace 1, \dots, n \rbrace$. 
The permutation importance of $\xi$ w.r.t.\ to the test set $\pmb{x}$ and permutation $\pi$ is defined by
\begin{eqnarray}
	\Delta \left(\xi; \pi \right) := \overline{\text{CRPS}} ( F(\tilde{\pmb{x}}^{\pi}_{\cdot}), y_{\cdot}) - \overline{\text{CRPS}} ( F(\pmb{x}_{\cdot}), y_{\cdot}),
	\label{eq:permuation_importance}
\end{eqnarray}
where $\tilde{\pmb{x}}^{\pi}$ is the data permuted in $\xi$ w.r.t.\ $\pi$, which is given by
\begin{eqnarray*}
\tilde{\pmb{x}}^{\pi}_{l,j} := \tilde{\pmb{x}}^{\pi}_{l,j}(\xi) := \begin{cases}
\pmb{x}_{l,j} & l \neq i, \, l = 1, \dots, p,  \\
\pmb{x}_{l, \pi(j)} & l = i,
\end{cases}
\qquad \text{for} \quad j = 1, \dots, n.
	\label{eq:permutation_single}
\end{eqnarray*}
In a nutshell, we shuffle the samples of $\xi$ in the test set, generate the forecasts based on the permuted set, calculate the associated CRPS and calculate the difference to the CRPS of the original data. The larger the difference, the more detrimental is the effect of shuffling the feature to the forecast performance, and thus the more important it is.

This procedure can also be applied on a set of $K \leq p$ features $\Xi$ corresponding to the indices $I = \lbrace i_1, \dots, i_K \rbrace$, which we refer to as multi-pass permutation importance \citep{McGovern2019}. In this case, we do not permute only one feature according to $\pi$ but instead the entire set $\Xi$, i.e.\
\begin{eqnarray*}
\tilde{\pmb{x}}^{\pi}_{l,j} := \tilde{\pmb{x}}^{\pi}_{l,j}(\Xi) = \begin{cases}
\pmb{x}_{l,j} & l \notin I, \, l = 1, \dots, p, \\
\pmb{x}_{l, \pi(j)} & l \in I,
\end{cases}
\quad \text{for} \quad j = 1, \dots, n.
	\label{eq:permutation_multiple}
\end{eqnarray*}
The permutation importance $\Delta (\Xi; \pi)$ is then calculated according to \eqref{eq:permuation_importance}. Finally, we calculate a \textit{relative permutation importance} via
\begin{eqnarray*}
	\Delta_0 \left(\xi; \pi \right) := \dfrac{ \Delta \left(\xi; \pi \right) }{ \overline{\text{CRPS}} ( F(\pmb{x}_{\cdot}), y_{\cdot}) }.
	\label{eq:relative_permuation_importance}
\end{eqnarray*}

\section{Implementation details} \label{sec:appendix_implementation}

We here provide additional details on the implementation of all methods. The corresponding \texttt{R} code is publicly available at \url{https://github.com/benediktschulz/paper_pp_wind_gusts}.

If the evaluation or parts of it are based on a set of quantiles, we generate 125 equidistant quantiles for each test sample. This number is chosen such that the median as well as the quantiles at the levels of a prediction interval with a nominal coverage corresponding to a 20-member ensemble (ca.\ 90.48\%) are included and such that the forecast distribution is given by a sufficiently large number of quantiles. 

\subsubsection*{Basic methods}

The implementation of EMOS and MBM is straightforward using built-in optimization routines in \texttt{R} in combination with functionalities for evaluation provided in the \texttt{scoringRules} package. We generally employed the gradient-based L-BFGS-B algorithm and in case of non-converged or failed optimizations applied the \code{Nelder-Mead} algorithm in a second run. In case of EMOS, we parameterized the parameter $b$ via $b = \exp ( \tilde{b} )$ with $\tilde{b} \in \R$, what resulted in a more stable estimation. Regarding the incorporation of the sub-ensemble structure into MBM, we further tested leaving out the parameter $d$, both with a single $c$ and $c_1, \dots, c_4$ for the sub-ensembles, but this resulted in a worse predictive performance still exhibiting systematic deviations in the histograms. The implementation of IDR relies on the \code{isodistrreg} package \citep{Henzi2019isodistrreg}, the associated hyperparameters are displayed in Table \ref{tbl:methods_config}.

\subsubsection*{EMOS-GB and QRF}

EMOS-GB and QRF are implemented via the \code{crch} \citep{Messner2016crch} and \code{ranger} \citep{Wright2017ranger} packages, respectively. The selected hyperparameter configurations are summarized in Table \ref{tbl:methods_config}. The permutation importance of the QRF predictor variables on the training set is obtained from the \code{ranger} function.

\subsubsection*{NN-based methods (DRN, BQN, HEN)}

Before fitting the NN models, each predictor variable was normalized by subtracting the mean value and dividing by the standard deviation based on the training set excluding the validation period. The NN models are built via the \code{keras} package \citep{Allaire2020keras}. For hyperparameter tuning, we used the \code{tfruns} package \citep{Allaire2018tfruns} to find the best hyperparameter candidates for a single run, for which we then compared an ensemble of ten corresponding network models. 
For pre-specified hyperparameters related to the basic model structures are provided in Table \ref{tbl:methods_config}, the hyperparameters selected with this procedure are displayed in Table \ref{tbl:methods_config_nn}. 

Except for HEN, where we employ the categorical cross-entropy, we manually implemented the loss functions for model training. For DRN, the CRPS of the truncated logistic distribution makes use of the \code{tf-probability} extension \citep[0.11.1;][]{Keydana2020tfprobability} that includes distribution functions of the logistic distribution. For BQN, the quantile loss evaluated at a given set of quantile levels based on a linear combination of Bernstein polynomials is implemented manually. 

For fitting the HEN model, the observations have to be transformed to categorical variables representing the bins, which are generated iteratively based on the training set. We start with one bin for each observation (which only take a certain amount of values for reporting reasons) and merge the bin that contains the least amount of observations with the smaller one of the neighbouring bins. We additionally put constrains on the bins. The first bin should have a length of at most 2 m/s, the last at most 7 m/s and the others at most 5 m/s. 

\begin{table}[p]
\caption{Final hyperparameter configurations of the methods presented in Section \ref{sec:methods}. The NN-specific hyperparameters are displayed separately in Table \ref{tbl:methods_config_nn}. \label{tbl:methods_config}}
\begin{center}
\begin{tabularx}{\linewidth}{@{}>{\bfseries}l@{\hspace{1em}}X@{\hspace{1em}}l@{\hspace{1em}}}
	\toprule 
	Method & Hyperparameter  & Value \\
	\midrule
	IDR & Number of subsamples & 100 \\
	& Subsample ratio & 0.5 \\
	& Maximum number of iterations (\code{max\_iter}) & 1,000 \\
	& Absolute threshold (\code{eps\_abs}) & 0.001 \\ 
	& Relative threshold (\code{eps\_rel}) & 0.001 \\
	\midrule
	EMOS-GB & Number of maximum iterations & 1,000 \\
	& Step size & 0.05 \\ 
	& Stopping criterion & AIC \\ 
	\midrule
	QRF & Number of trees & 1,000 \\
	& Ratio of predictors considered at each split & 0.5 \\ 
	& Minimal node size & 5 \\
	& Maximal tree depth & 20 \\
	\midrule
	BQN & Degree of Bernstein polynomials & 12 \\
	& Number of equidistant quantiles in quantile loss & 99 \\
	\midrule
	HEN & Number of (non-equidistant) bins & 20  \\
	\bottomrule
\end{tabularx}
\end{center}

\bigbreak\bigbreak

\caption{Overview of the configuration of the individual networks in the NN-based methods. \label{tbl:methods_config_nn}}	
\begin{center}
\begin{tabularx}{\linewidth}{X@{\hspace{2em}}c@{\hspace{3em}}c@{\hspace{3em}}c@{\hspace{3em}}}
	\toprule 
	Hyperparameter & DRN & BQN & HEN  \\
	\midrule
	Learning rate & $5 \cdot 10^{-4}$ & $5 \cdot 10^{-4}$ & $5 \cdot 10^{-4}$ \\ 
	Epochs & 150 & 150 & 150 \\ 
	Patience & 10 & 10 & 10 \\ 
	Batch size & 64 & 64 & 64 \\ 
	Embedding dimension & 10 & 10 & 10 \\ 
	Hidden layers & 2 & 2 & 2 \\ 
	Nodes per layer & (64, 32) & (48, 24) & (64, 32) \\ 
	Activation & softplus & softplus & softplus \\ 
	Output nodes & 2 & 13 & 20 \\ 
	Output activation & softplus & softplus & softmax \\ 
	\midrule
	Size of network ensemble & 10 & 10 & 10 \\ 
	\bottomrule
\end{tabularx}
\end{center}
\end{table}

\clearpage
\newpage

%
%

\setcounter{figure}{0}

\makeatletter 
\renewcommand{\thefigure}{S\@arabic\c@figure}
\makeatother

\setcounter{section}{0}
\renewcommand{\thesection}{S\arabic{section}}

\section*{Supplemental material}

\section{Additional figures}

\subsection{Additional illustrations for MBM}

\begin{figure}[h!]\begin{center}
		\scalebox{0.2}{\includegraphics{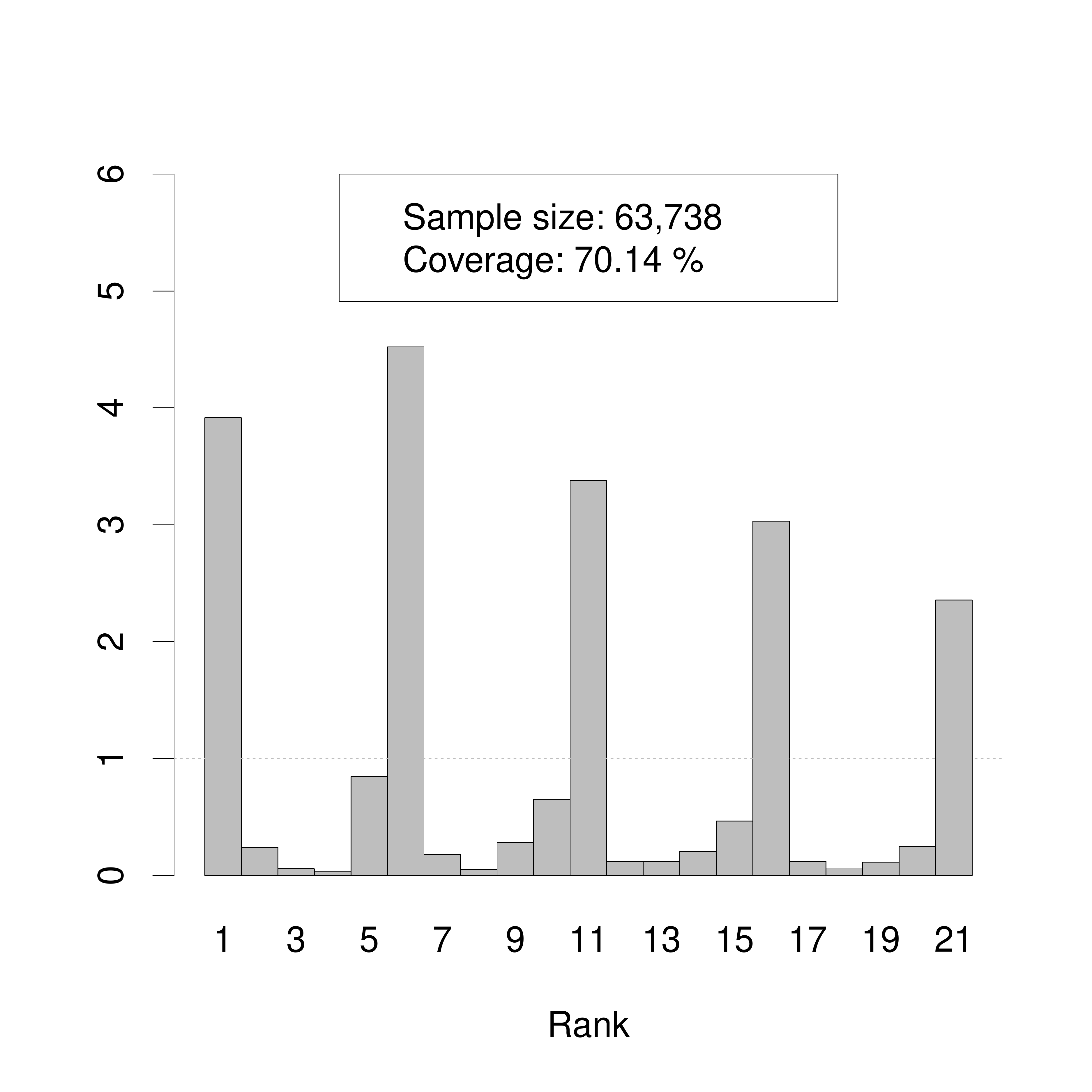}}
		\scalebox{0.2}{\includegraphics{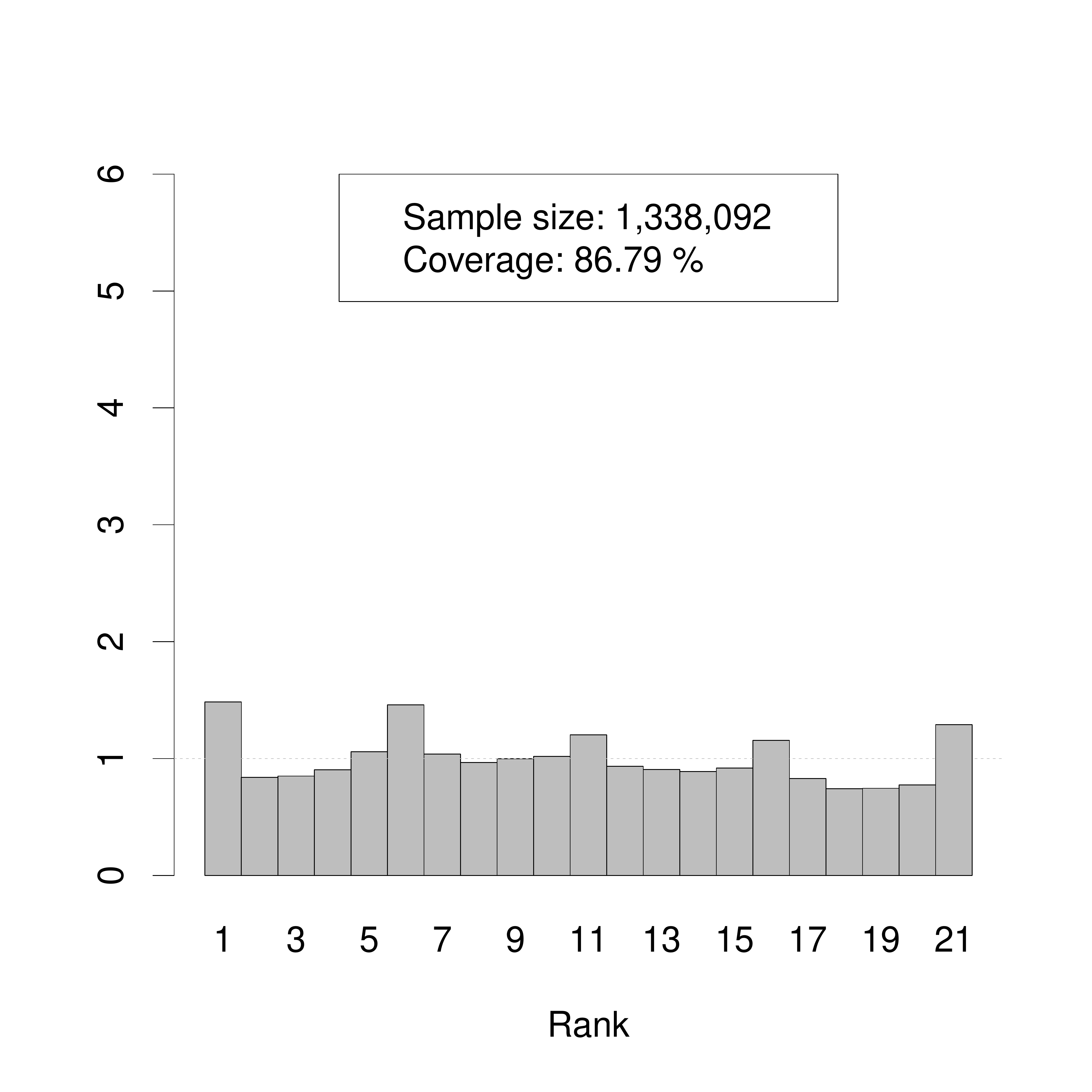}}
		\caption{Verification rank histograms of MBM forecasts based on \eqref{eq:mbm_general} with CRPS estimation on the entire training set for a lead time of 0h (left) and greater than 0h (right). Coverage refers to the empirical coverage of a prediction interval with a nominal coverage corresponding to a 20-member ensemble (ca.\ 90.48\%). \label{supl_fig:mbm_hist_incorp}}
\end{center}\end{figure}

\begin{figure}[h]\begin{center}
		\includegraphics[width=\textwidth]{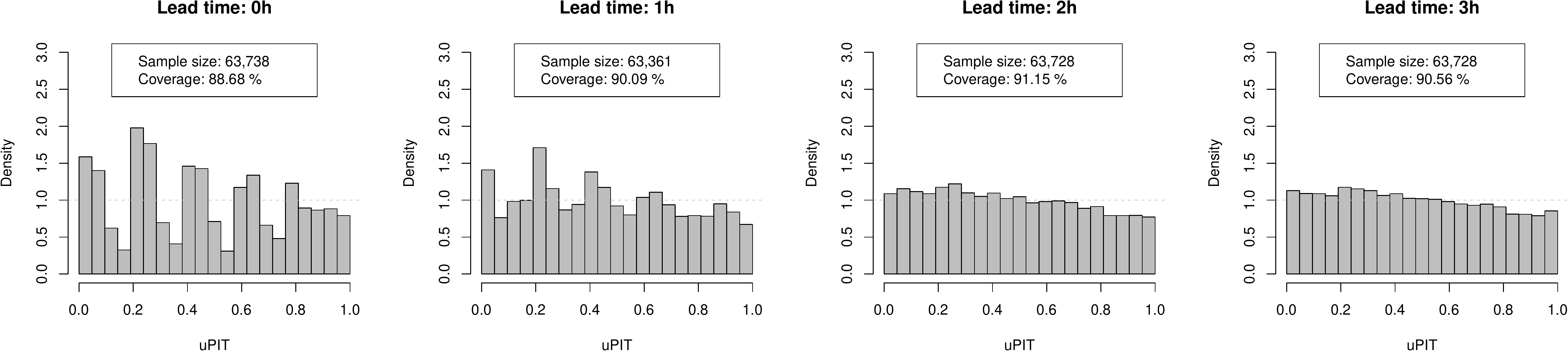}
		\caption{uPIT histograms of the MBM forecasts for lead times from 0 to 3h (left to right) over all stations. Coverage refers to the empirical coverage of a prediction interval with a nominal coverage corresponding to a 20-member ensemble (ca.\ 90.48\%). \label{supl_fig:eval_mbm_histograms_steps0_3}}
\end{center}\end{figure}


\newpage 

\subsection{Climatological differences of the training and test dataset}

\begin{figure}[h!]
\begin{center}
\includegraphics[width=\textwidth]{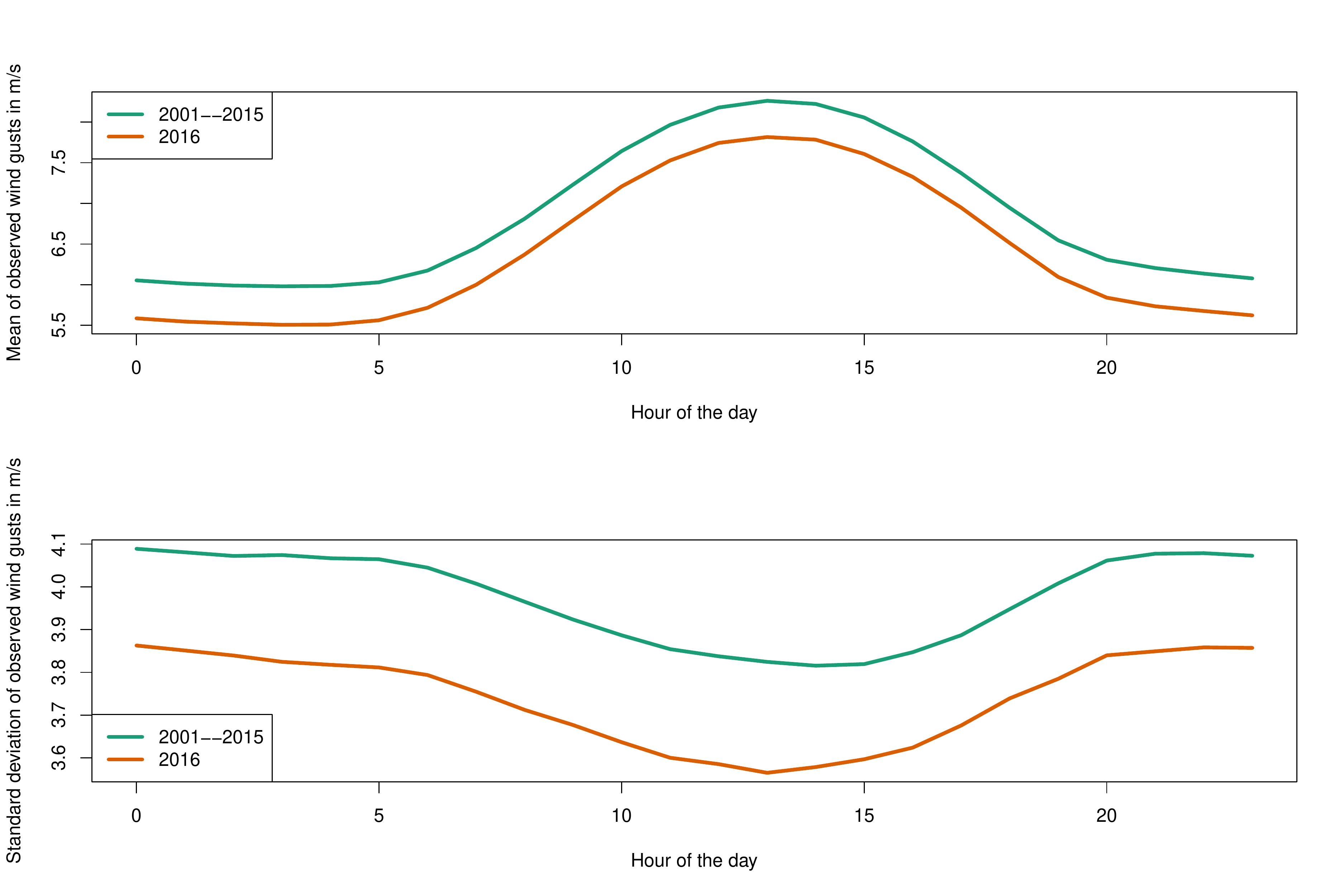}
\caption{Mean (left) and standard deviation (right) of the observed wind gusts dependent on the hour of the day in the years 2001--2015 (green) and 2016 (orange). \label{supl_fig:eval_epc_mean_wind_gusts}}
\end{center}
\end{figure}

\newpage

\subsection{Additional feature importance results}

\subsubsection{EMOS-GB}

\begin{figure}[h]
\begin{center}
\includegraphics[width=\textwidth]{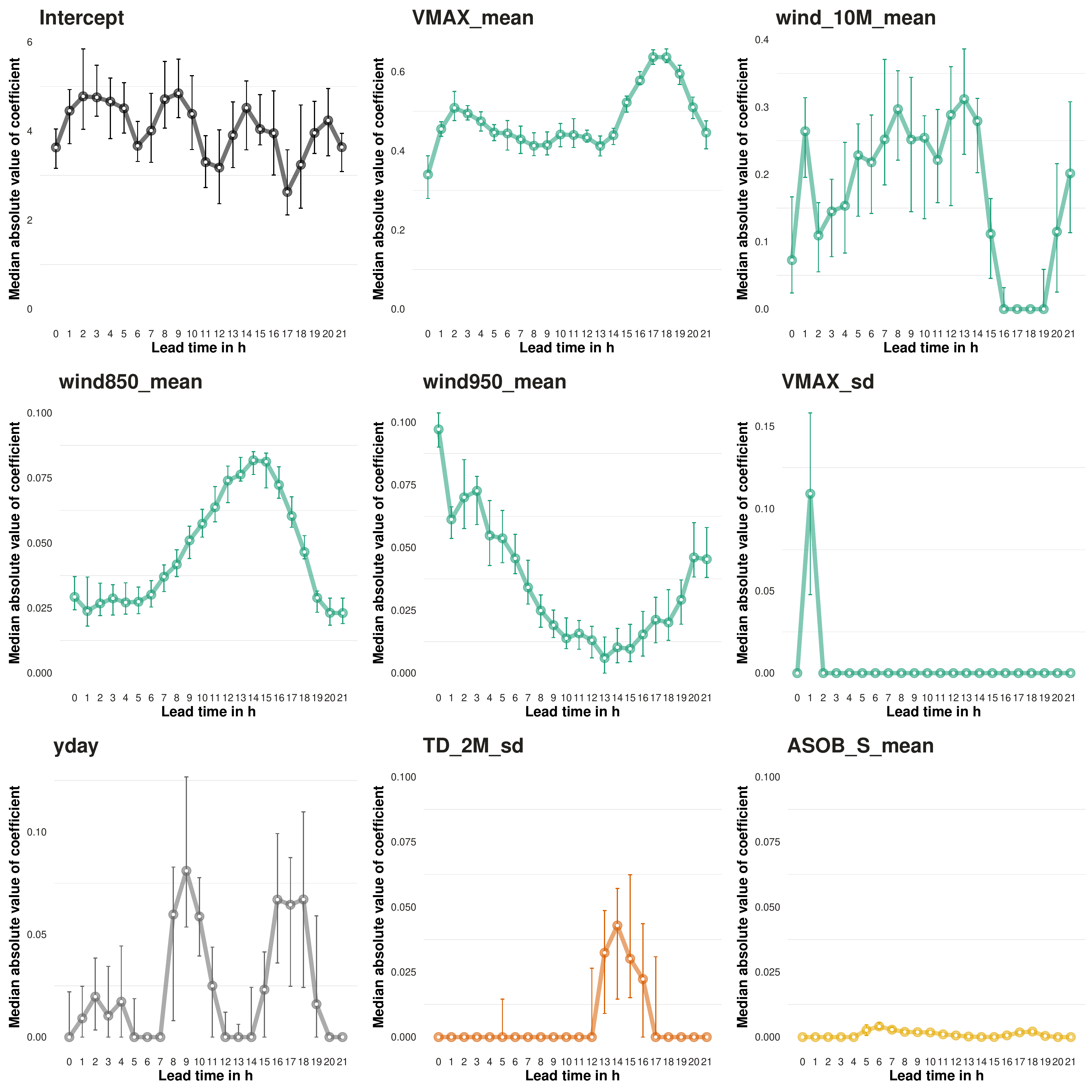}
\caption{Median of station-wise absolute values of the location parameter coefficients for selected predictors (see Table \ref{tbl:data_predictors}) of the EMOS-GB model as functions of the forecast lead time. The error bars indicate a bootstrapped 95\%-confidence interval of the median. Note the different scale of the vertical axes. 
\label{supl_fig:fi_emos_gb_loc}}
\end{center}
\end{figure}

\begin{figure}[p]
\begin{center}
\includegraphics[width=\textwidth]{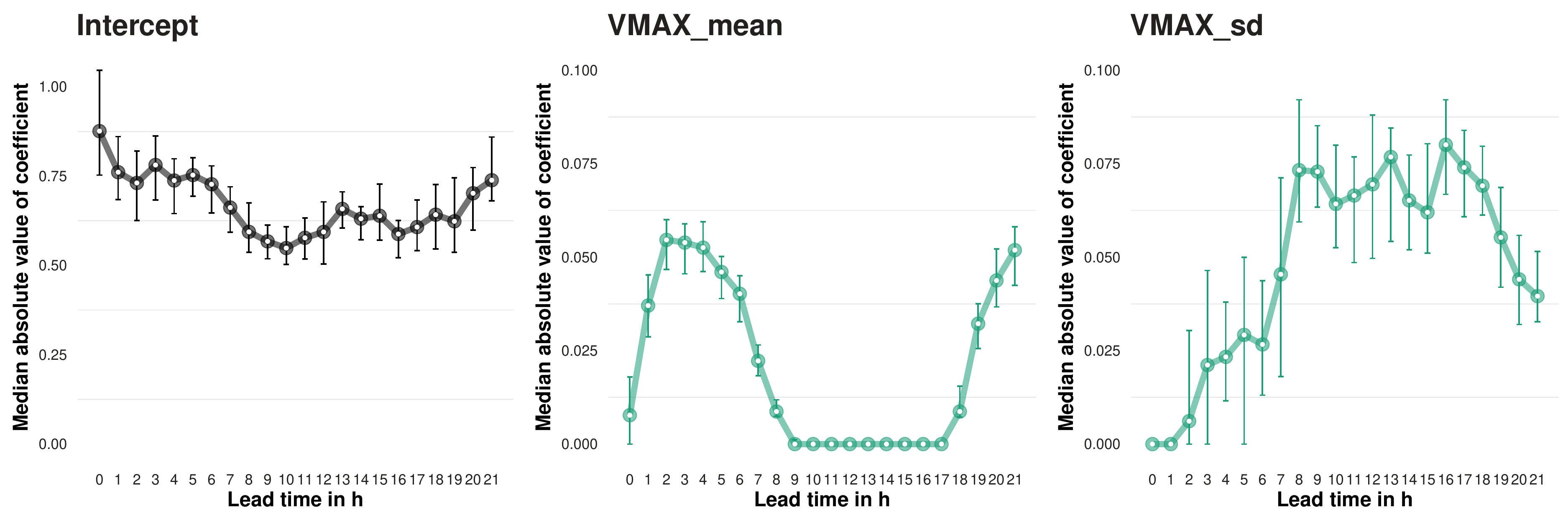}
\caption{Median of station-wise absolute values of the scale parameter coefficients for selected predictors (see Table \ref{tbl:data_predictors}) of the EMOS-GB model as functions of the forecast lead time. The error bars indicate a bootstrapped 95\%-confidence interval of the median. Note the different scale of the vertical axes. 
\label{supl_fig:fi_emos_gb_scale}}
\end{center}

\bigbreak
\bigbreak 

\begin{center}
\includegraphics[width=\textwidth]{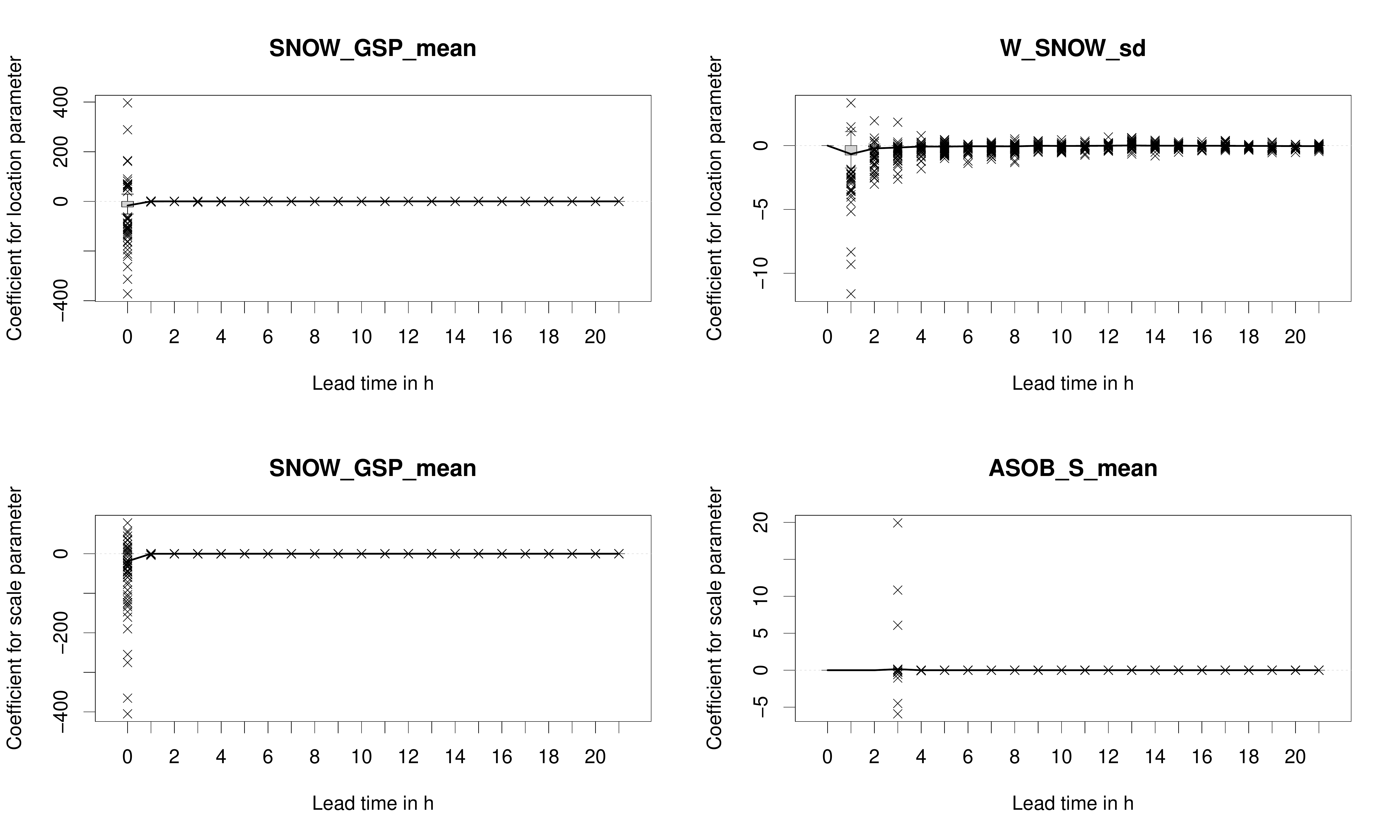}
\caption{Boxplots of station-wise parameter coefficients for selected predictors (see Table \ref{tbl:data_predictors}) of the EMOS-GB model as functions of the forecast lead time, illustrating severe outliers for some combinations of stations and lead times. The solid line indicates the mean coefficient values averaged over all stations. Note the different scale of the vertical axes.
\label{supl_fig:fi_emos_gb_outliers}}
\end{center}
\end{figure}

\clearpage  

\subsubsection{QRF}

\begin{figure}[h]
\begin{center}
\includegraphics[width=\textwidth]{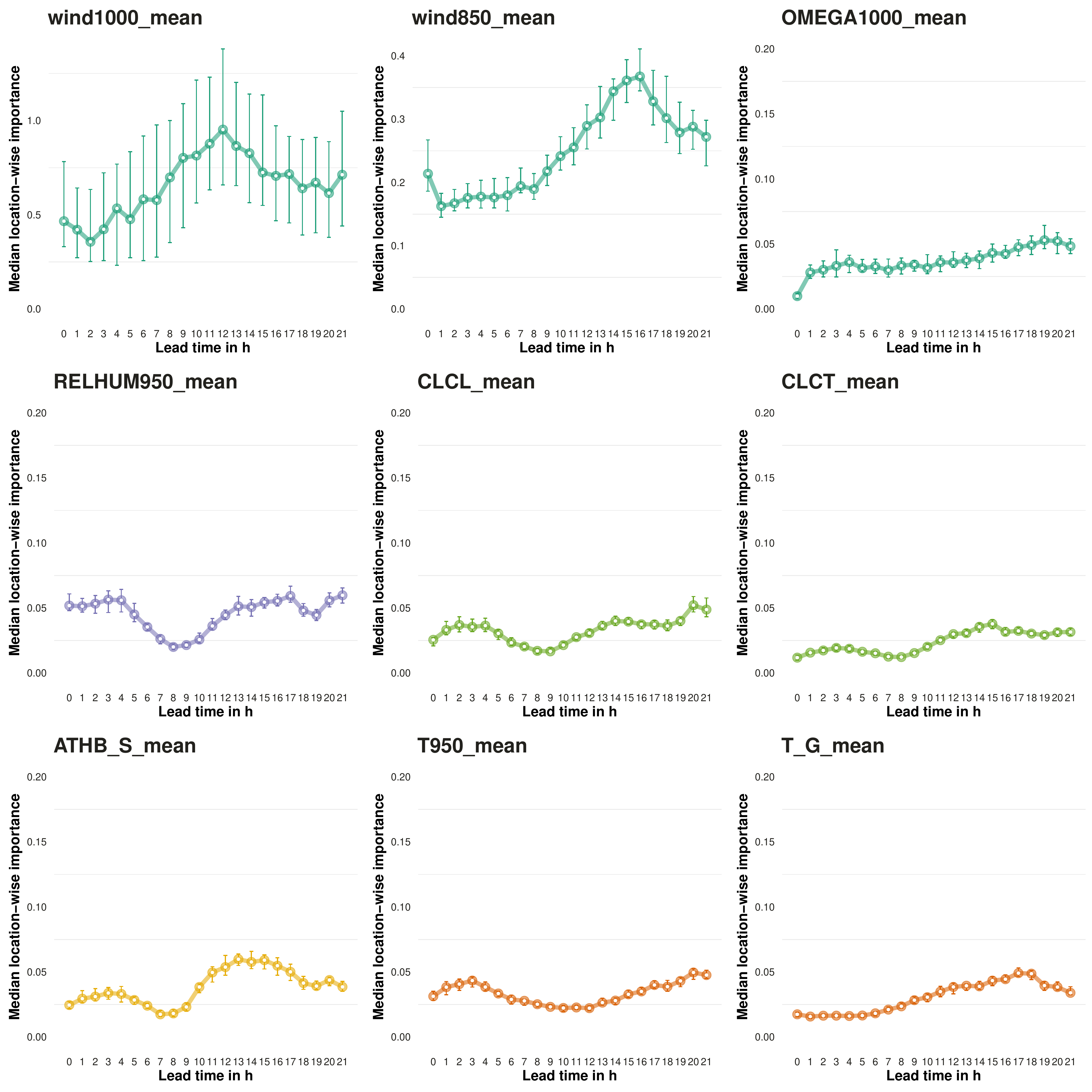}
\caption{Median of station-wise feature importance for selected predictors (see Table \ref{tbl:data_predictors}) of the QRF model as functions of the forecast lead time. The error bars indicate a bootstrapped 95\%-confidence interval of the median. Note the different scale of the vertical axes. 
\label{supl_fig:fi_qrf}}
\end{center}
\end{figure}

\clearpage 

\subsubsection{NN models}

\begin{figure}[h]
\begin{center}
\includegraphics[width=\textwidth]{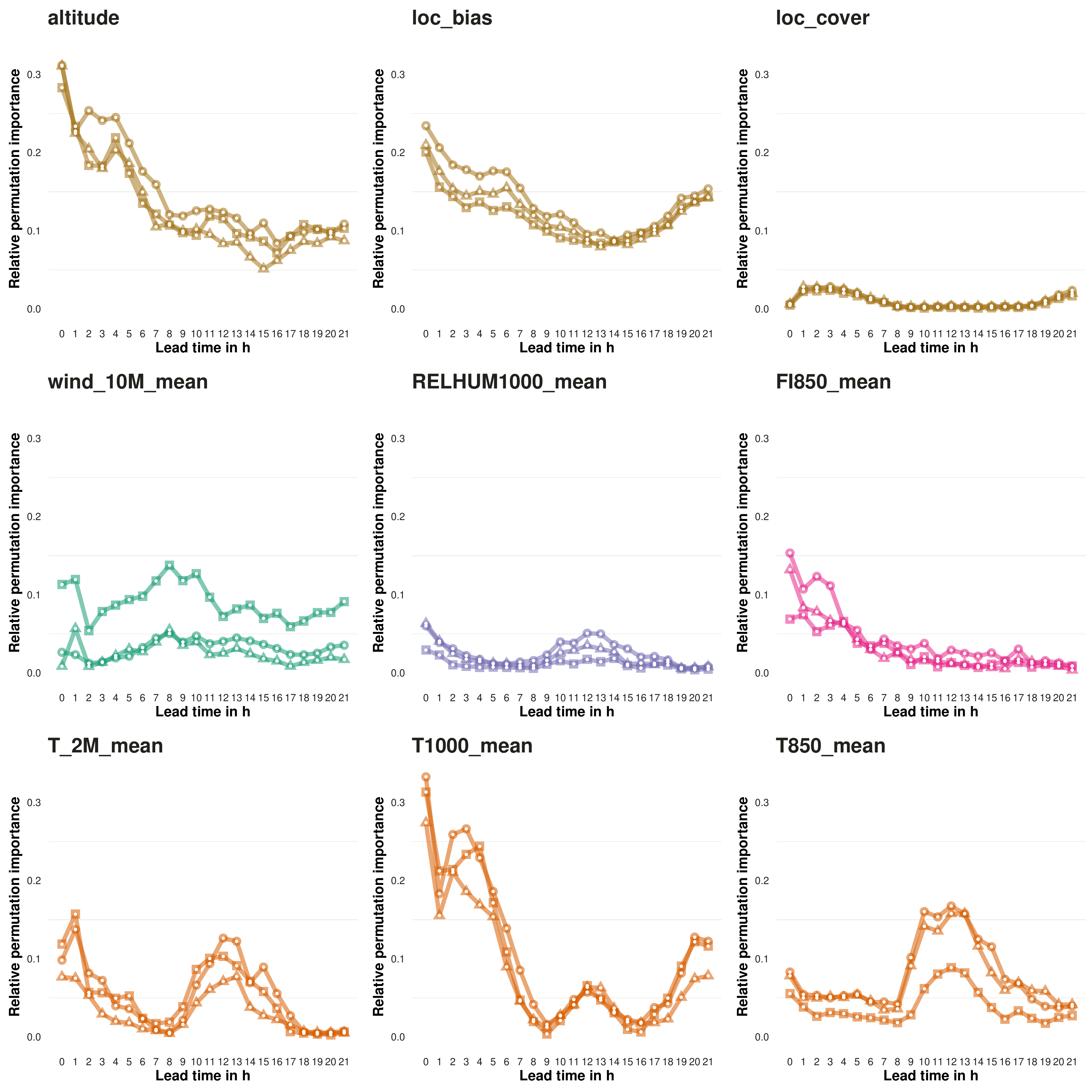}
\caption{Relative permutation importance of selected predictors (see Table \ref{tbl:data_predictors}) for the three NN-based model variants dependent on the lead time. Note the different scale of the vertical axes.
Different symbols indicate the three model variants ($\circ$: DRN, $\triangle$: BQN, $\square$: HEN). \label{supl_fig:eval_fi_nn_features}}
\end{center}
\end{figure}

\begin{figure}
\begin{center}
		\includegraphics[width=\textwidth]{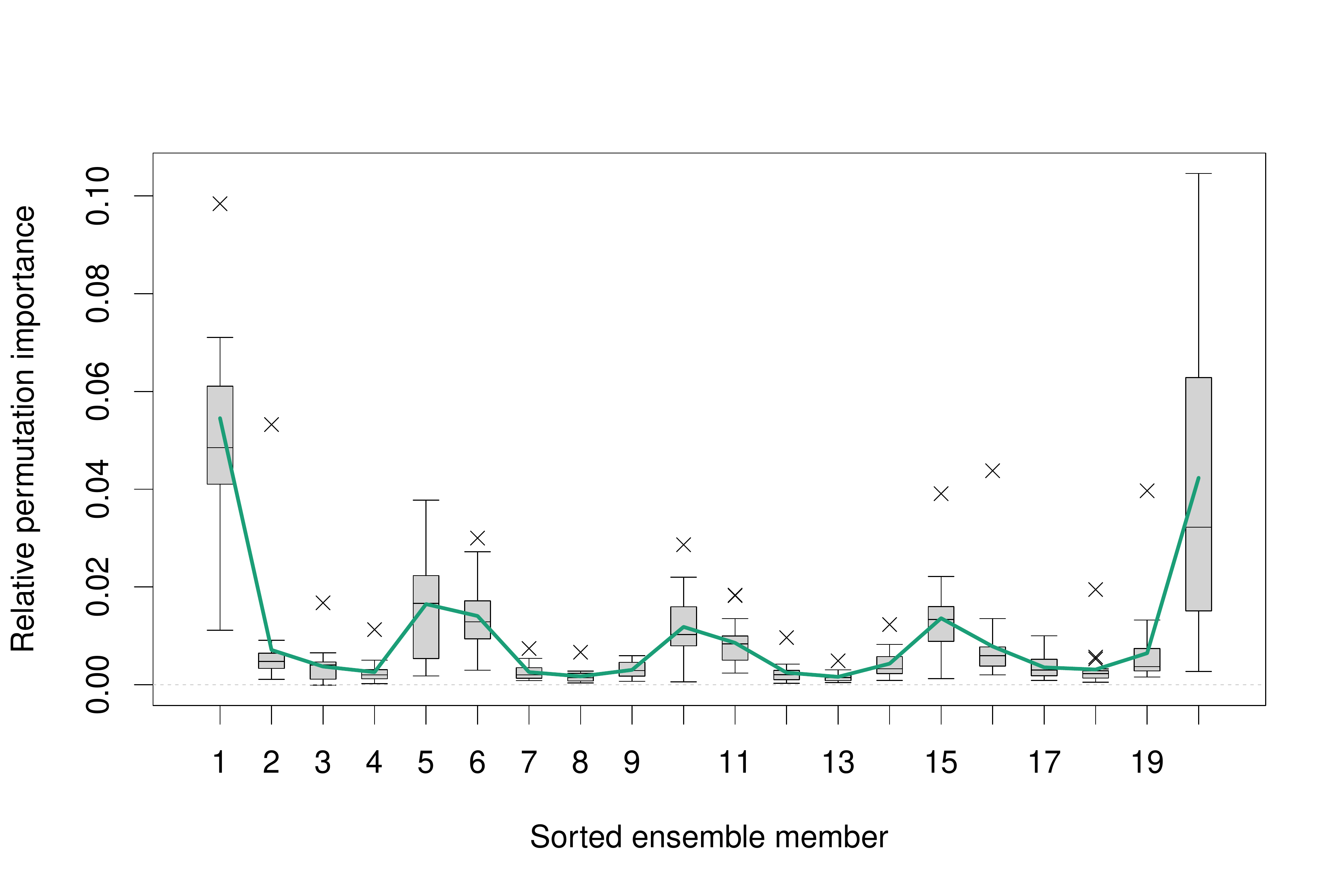}
		\caption{Boxplots of the relative permutation importance of the sorted ensemble members in case of BQN.
		\label{supl_fig:eval_fi_bqn_members}}
\end{center}\end{figure}

\clearpage

\section{CRPS of a piecewise (uniform) distribution} \label{sec:supl_crps_hen}

Let $y \in \R$, $N$ be the number of bins and $b_0 < \dots < b_N$ the edges of the bins $I_l = [b_{l-1}, b_l)$ with probabilities $p_l$, $l = 1, \dots, N$, where it holds that $\sum_{l = 1}^{N} p_l = 1$. Then, the CRPS of a piecewise uniform distribution is given by
\begin{eqnarray}
\text{CRPS} (F, y) = | y - \tilde{\tilde{y}} | + \sum_{l = 1}^{N} \text{CRPS}  \left( G_{a_l, L}^{b_{l-1}, U}, \tilde{y}_l \right), \label{eq:hen_crps}
\end{eqnarray}
where $ \tilde{\tilde{y}} = \min \left( b_N,  \max \left( b_0, y \right)\right)$, $ \tilde{y}_l = \min \left( b_l,  \max \left( b_{l-1}, y \right)\right)$ for $l = 1, \dots, N$ and $G_{b_{k-1}, L}^{b_k, U}$ denotes a uniform distribution with lower edge $b_{k-1}$, upper edge $b_k$, point mass $L := \sum_{l = 1}^{k - 1} p_l$ on the lower edge and point mass $U := 1 - \sum_{l = k + 1}^{N} p_l$ on the upper edge. Note that the first term penalizes the case that the observation falls out of the range of the bins. The CRPS of a uniform distribution with point masses on the edges is given in \cite{Jordan2019scoringrules}.

For the proof, we state the CRPS of a piecewise distribution that can be calculated directly via the CRPS of the individual distributions. 
A variant of this result, which we adapt to our setting and notation, is given in \citet{Jordan2016}, and a proof is provided below.
The HEN forecast is included as a special case, since the CDF \eqref{eq:hen_cdf} of the HEN forecast is equivalent to that of a uniform distribution with lower edge $b_{k-1}$, upper edge $b_k$, point mass $L := \sum_{l = 1}^{k - 1} p_l$ on the lower edge and point mass $U := 1 - \sum_{l = k + 1}^{N} p_l$ on the upper edge for a fixed $k$.

\subsubsection*{To show:}

Let $G$ denote a continuous CDF defined by the piecewise PDF
$$ g(z) = \sum_{k = 1}^{N} g_k (z) \mathbbm{1} \lbrace z \in I_k \rbrace, \quad z \in \R.$$
For $k = 1, \dots, N$, define $G_k$ as the CDF defined by $g_k$ that has support $\bar{I_k}$ and point masses
$$ L = \int_{-\infty}^{b_{k-1}} g(z) dz \quad \text{resp.} \quad U = \int_{b_k}^{\infty} g(z) dz $$
on $b_{k-1}$ resp.\ $b_k$ and define $\tilde{y}_k := \min \lbrace b_k, \max \lbrace b_{k-1}, y \rbrace \rbrace$. Then, it holds that
$$ \text{CRPS} (G, y) = | y - \tilde{\tilde{y}} | + \sum_{k = 1}^{N} \text{CRPS}  \left( G_k, \tilde{y}_k \right), \quad \tilde{\tilde{y}} := \min \lbrace b_N, \max \lbrace b_0, y \rbrace \rbrace. $$

\subsubsection*{Proof:}

Define $h \left(z; y, F\right) := \left( F (z) - \mathbbm{1} \lbrace y \leq z \rbrace \right)^2$ for $z, y \in \R$. First, we can note for all $z \notin \text{supp}(F)$, where $l$ and $u$ are the lower and upper edges of the support of $F$, that
\begin{eqnarray}
h \left(z; y, F\right) = \left\lbrace \begin{array}{lll}
1, & \text{if} \quad y \leq z < l, & \quad (i) \\
0, & \text{if} \quad z < l \quad \text{and} \quad z < y, & \quad (ii) \\
0, & \text{if} \quad l \leq y < u, & \quad (iii) \\
0, & \text{if} \quad u \leq z \quad \text{and} \quad y \leq z, & \quad (iv) \\
1, & \text{if} \quad u \leq z < y. & \quad (v)
\end{array}\right. \nonumber
\end{eqnarray}
Let $y \in \R$. Then
\begin{eqnarray}
\text{CRPS} (G, y) &=& \int_{\R} h \left( z; y, G \right) dz \nonumber \\[1em] 
&=& \int_{- \infty}^{b_0} h \left( z; y, G \right) dz + \int_{b_N}^{\infty} h \left( z; y, G \right) dz + \sum_{k = 1}^{N} \int_{I_k} h \left( z; y, G \right) dz \nonumber \\[1em] 
&\overset{(I)}{=}& | y - \tilde{\tilde{y}} | + \sum_{k = 1}^{N} \int_{I_k} h \left( z; y, G_k \right) dz \nonumber \\[1em] 
&\overset{(II)}{=}& | y - \tilde{\tilde{y}} | + \sum_{k = 1}^{N} \int_{I_k} h \left( z; \tilde{y}_k, G_k \right) dz \nonumber \\[1em] 
&\overset{(III)}{=}& | y - \tilde{\tilde{y}} | + \sum_{k = 1}^{N} \int_{\R} h \left( z; \tilde{y}_k, G_k \right) dz \nonumber \\[1em] 
&=& | y - \tilde{\tilde{y}} | + \sum_{k = 1}^{N} \text{CRPS}  \left( G_k, \tilde{y}_k \right). \nonumber
\end{eqnarray}
$\underline{(I)}$:
\begin{enumerate}
	\item If $y \leq b_0$, we have $ \int_{b_N}^{\infty} h \left( z; y, G \right) dz \overset{(iv)}{=} 0 $
	$$ \int_{- \infty}^{b_0} h \left( z; y, G \right) dz = \int_{- \infty}^{y} h \left( z; y, G \right) dz + \int_{y}^{b_0} h \left( z; y, G \right) dz \overset{(ii) \& (i)}{=} 0 + | y - \tilde{\tilde{y}} |. $$
	\item If $y \geq b_N$, we have $ \int_{- \infty}^{b_0} h \left( z; y, G \right) dz \overset{(ii)}{=} 0 $
	$$ \int_{b_N}^{\infty} h \left( z; y, G \right) dz = \int_{b_N}^{y} h \left( z; y, G \right) dz + \int_{y}^{\infty} h \left( z; y, G \right) dz \overset{(v) \& (iv)}{=} | y - \tilde{\tilde{y}} | + 0. $$
\end{enumerate}
$\underline{(II)}$:
Let $k \in \lbrace 1, \dots, N \rbrace$. We need show that $ h \left( z; \tilde{y}_k, G_k \right) = h \left( z; y, G_k \right) $ for all $ z \in I_k $. This holds if $ \mathbbm{1} \lbrace \tilde{y}_k \leq z \rbrace = \mathbbm{1} \lbrace y \leq z \rbrace $ for all $z \in I_k$, what follows from:
\begin{enumerate}
	\item $\tilde{y}_k = b_{k-1}$: Then $y \leq \tilde{y}_k = b_{k-1} < z$ for all $ z \in I_k$, so $ \mathbbm{1} \lbrace \tilde{y}_k \leq z \rbrace = \mathbbm{1} \lbrace y \leq z \rbrace = 1 $.
	\item $\tilde{y}_k = y$: $\checkmark$.
	\item $\tilde{y}_k = b_k$: Then $y \geq \tilde{y}_k = b_k > z$ for all $ z \in I_k$, so $ \mathbbm{1} \lbrace \tilde{y}_k \leq z \rbrace = \mathbbm{1} \lbrace y \leq z \rbrace = 0 $.
\end{enumerate}
$\underline{(III)}$:
\\
Let $k \in \lbrace 1, \dots, N \rbrace$. Property $(iii)$ implies that $ h \left( z; \tilde{y}_k, G_k \right) = 0$ for $z \in \R \backslash I_k$ and thus
$$ \int_{\R} h \left( z;  \tilde{y}_k, G_k \right) dz = \int_{I} h \left( z;  \tilde{y}_k, G_k \right) dz. $$
\hfill $\qed$

\end{document}